\documentclass[10pt,a4paper,fleqn]{article}

\usepackage[
    top=20mm,
    bottom=20mm,
    left=18mm,
    right=18mm
]{geometry}

\usepackage{graphicx}
\usepackage{multicol}
\usepackage{titlesec}
\usepackage{fancyhdr}
\usepackage{setspace}
\usepackage{array}
\usepackage{booktabs}
\usepackage{tikz}
\usepackage{tcolorbox}
\usepackage{amssymb}
\usepackage{caption}
\usepackage{fontspec}
\usepackage{amsmath}
\usepackage{graphicx}
\usepackage{siunitx}
\usepackage{booktabs} 
\usepackage{float}
\usepackage{makecell}
\usepackage{multirow}
\usepackage{comment}

\usepackage{hyperref}
\usepackage{url}
\usepackage{lipsum}

\usepackage{booktabs}
\usepackage{titlesec}
\titleformat{\section}
    {\normalsize\bfseries}
    {\thesection}
    {1em}
    {}

\titleformat{\subsection}
  {\footnotesize\bfseries}
  {\thesubsection}
  {1em}
  {}

\titleformat{\subsubsection}
  {\footnotesize\bfseries}
  {\thesubsubsection}
  {1em}
  {}

\titleformat{\paragraph}[runin]
  {\footnotesize\bfseries}
  {\theparagraph}
  {1em}
  {}

\usepackage{lipsum}

\usepackage{varwidth}
\usepackage{adjustbox}

\definecolor{tether-gray}{HTML}{F2F1EF}

\newtcolorbox[auto counter]{custombox}[2][]{
    title={Box~\thetcbcounter: #2},
    label={#1},
    fonttitle=\bfseries,
    coltitle=black,
    before title=\vspace{15pt},
    colback=tether-gray,
    colframe=tether-gray,
    boxrule=0pt,
    arc=5pt,
    left=15pt,
    right=15pt,
    bottom=15pt,
    width=\textwidth,    
}

\newcommand{\tablebox}[1]{
    \tcbox[
        colback=tether-gray,
        colframe=tether-gray,
        boxrule=0pt,
        left=15pt,
        right=15pt,
        top=15pt,
        bottom=15pt
    ]{#1}
}

\begin{document}


\begin{flushleft}
\includegraphics[height=20pt]{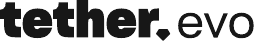}
\end{flushleft}

\vspace{10pt}


\begin{tcolorbox}[
    colback=tether-gray,
    colframe=tether-gray,
    boxrule=0pt,
    arc=5pt,
    left=15pt,
    right=15pt,
    top=15pt,
    bottom=15pt,
    width=\textwidth
]


{\huge\bfseries
REEL: Brain-Guided Video Generation and Selection by Reranking Encoder Embeddings with Linear Decoding from fMRI data
}

\vspace{5pt}


{\small\bfseries
Fabrizio Spera\textsuperscript{1,\dag},
Tommaso Boccato\textsuperscript{2},
Moein Khajehnejad\textsuperscript{2,3},
Nicola Toschi\textsuperscript{1,4}\textsuperscript{*},
Matteo Ferrante\textsuperscript{1,2}\textsuperscript{*}
}

\vspace{2.5pt}

{\small
\textsuperscript{1}Department of Biomedicine and Prevention, University of Rome Tor Vergata, Rome, Italy\\
\textsuperscript{2}Tether Evo\\
\textsuperscript{3}Monash University, Melbourne, Australia.\\
\textsuperscript{4}A.A.
Martinos Center for Biomedical Imaging, Massachusetts General Hospital, Boston, USA\\
\textsuperscript{*} Equal Contribution\\
\textsuperscript{\dag} Correspondence : Fabrizio.Spera@uniroma2.it
}

\vspace{15pt}


{\footnotesize
Visual brain decoding from functional magnetic resonance imaging (fMRI) data has made remarkable progress in recent years, driven by increasingly powerful artificial intelligence (AI) models and the growing availability of large-scale public datasets. However, despite the impressive advances achieved in static image decoding, the reconstruction of videos from fMRI signals remains challenging due to the temporal mismatch between rapidly changing video stimuli and the slow dynamics of the blood-oxygen-level-dependent (BOLD) response. We evaluate a Bayesian-inspired proposal-and-rerank strategy on CC2017 and the BOLD Moments Dataset (BMD), using a linear semantic decoder, a pretrained video generator, and three neural encoding streams with ridge readouts and learned pooling in two streams. We refer to this framework as REEL (\textit{R}eranking \textit{E}ncoder \textit{E}mbeddings with \textit{L}inear decoding). Compared with a single generated candidate, reranking increases mean Video 50-way accuracy from $0.2359$ to $0.2434$ across three CC2017 participants and from $0.3653$ to $0.3920$ across ten BMD participants. These findings suggest an alternative and conceptually simpler strategy for improving video brain decoding, highlighting the potential of our approach as a complementary direction to increasingly complex pipelines.
}

\end{tcolorbox}

\vspace{10pt}


\footnotesize


\section{Introduction}
\label{sec:intro}
Decoding perceptual content from brain activity has long been a goal at the intersection of neuroscience and AI; fMRI, which offers non-invasive access to activity across the whole visual cortex, has enabled remarkable progress in this direction, driven both by the release of large-scale public datasets \cite{allen_massive_2022, bold5000, cc2017, Bold} and by increasingly powerful pretrained generative and representation models \cite{clip, latentdiffusionmodels, versatilediffusion}. Most of this progress, however, has concerned static images \cite{ozcelik_natural_2023, mindeye1,mindeye2, dynadiff}, whereas natural vision is inherently dynamic and unfolds as a continuous stream of scenes, objects and movements. Extending brain decoding from images to videos is therefore a natural step, but a considerably harder one, mainly because of the mismatch between the timescale of visual events and that of the fMRI signal: while the content of a video changes over tens of milliseconds, the BOLD response only reflects brain activity integrated over seconds, so that each measurement captures a temporally blurred summary of the stimulus rather than any individual frame. Thus, recovering semantic, spatial and dynamic information from such a signal is a far more complicated problem than in the static setting.\\
Existing fMRI-to-video pipelines have addressed this problem by decoding from brain activity a growing number of intermediate representations and using them jointly to condition a pretrained video generator \cite{mind-video,  brainnetflix, NeuroClips, NEURONS, mind2word, semvideo}. Although these multi-stage pipelines have improved reconstruction quality, each additional stage brings its own trainable components and design choices, making it difficult both to evaluate the contribution of each single component and to assess to what extent a reconstruction is actually supported by the neural signal, rather than by the other modules.\\
Through our pipeline REEL, we follow a different and conceptually simpler strategy, rooted in the Bayesian view of decoding, according to which a good reconstruction should be both plausible under a prior over natural stimuli and consistent with the measured brain response, as predicted by an encoding model. \cite{Nishimoto2011} applied this idea to natural movies by scoring a large set of video clips, and \cite{ciferri_bayesian} more recently showed, in the context of music reconstruction, that such a large prior set can be replaced by a generative model conditioned on brain-decoded embeddings, from which candidate reconstructions are sampled and then selected according to their consistency with the neural response. Our work builds directly on this prior-guided strategy and extends it to video, using the brain response twice: first to decide \emph{where to search} within the generative space, by decoding the conditioning of the generator, and then to decide \emph{which candidate to select}, by comparing predicted and measured activity. Transferring this approach to video is, however, not straightforward, since modern text-to-video generators are conditioned on high-dimensional token sequences and produce candidates that may share the same semantic content while differing in spatial layout and motion, so that the encoding models used for selection must capture low-level and dynamic information in addition to semantics.\\
Through a linear ridge decoder we map fMRI activity into the text-conditioning space of CogVideoX-5b \cite{cogvideo}, thus defining a brain-conditioned proposal from which we sample a set of candidate videos; three encoding models, targeting early, ventral and dorsal visual cortex voxels, then predict the neural response to each candidate, retaining as final reconstruction the candidate whose predicted activity is the most similar to the measured one.\\
We evaluate REEL on two public datasets with complementary designs: CC2017 \cite{cc2017}, in which clips are cut from long continuous movies, and the BMD \cite{Bold}, in which each 3-s clip depicts a self-contained event. With a single linearly decoded conditioning signal and encoding-based reranking, REEL achieves the highest reported SSIM among the CC2017 methods in Table \ref{tab:comparison}, while its semantic identification accuracies remain below those of SemVideo. On BMD, the reported scores are promising, although differences in participants, training regimes, generators, and evaluation protocols limit direct comparisons. The clearest controlled evidence is the improvement from reranking candidates generated by the same proposal.

\section{Related Work}
\label{sec:related}
The rapid development of diffusion models \cite{latentdiffusionmodels,versatilediffusion,videolatentdiffusion} has substantially advanced video generation. Building on image-generation architectures, video diffusion models introduce spatiotemporal components that capture motion and preserve temporal consistency across frames, and more recent architectures based on Diffusion Transformers (DiTs) \cite{dits} have further improved generation quality and long-range temporal coherence.
Text-to-video frameworks such as AnimateDiff \cite{animatediff} and CogVideoX \cite{cogvideo} now generate videos of remarkable quality, which makes them well suited as decoders for fMRI-based reconstruction. Their output, however, can only be as informative as the conditioning they receive, so that the quality of a reconstruction depends strongly on how accurately and richly this conditioning can be predicted from brain activity.\\
For this reason, most fMRI-to-video methods focus on mapping neural activity into the embedding spaces through which pretrained generators are conditioned. Depending on the generator, these spaces may be defined by CLIP embeddings \cite{clip}, as commonly done with AnimateDiff, or by text embeddings from language models such as T5 \cite{T5}, as in CogVideoX. Projecting noisy, high-dimensional neural signals into such representation spaces allows generation to be guided by knowledge learned from large-scale visual and textual data, and recent methods have progressively enriched this conditioning with further decoded signals, ranging from keyframes and blurry reconstructions to scene descriptions and recognized concepts.\\
On CC2017, MinD-Video \cite{mind-video} improves the semantic quality of the reconstructions through dedicated spatiotemporal attention modules, while NeuroClips \cite{NeuroClips} combines a semantic reconstructor with a perception reconstructor, the latter exploiting a pretrained fMRI-to-image model \cite{mindeye2} to provide an initial keyframe. NEURONS \cite{NEURONS} draws instead on the functional specialization of the visual cortex and decomposes video decoding into four complementary subtasks, namely key-object segmentation, concept recognition, scene description and blurry video reconstruction, whose outputs jointly guide generation. More recently, SemVideo \cite{semvideo} relies on a multimodal large language model to describe each training clip at three levels of textual supervision, whose CLIP text embeddings are predicted from fMRI by a cross-subject Semantic Alignment Decoder; a Motion Adaptation Decoder, guided by the decoded motion semantics through spatial, temporal and cross-attention, then predicts frame-wise VAE latents from which a coarse video is obtained.\\
Methods evaluated on BMD have followed a similar trajectory. Brain Netflix \cite{brainnetflix} adopts a three-stage pipeline trained across multiple subjects and datasets, in which an fMRI encoder is first pretrained with masked brain modeling on large-scale resting-state and task data and then contrastively aligned with the CLIP text embeddings of video captions; MLP regressors subsequently map the aligned representation both to an initial video latent and to a BLIP embedding that is decoded into a caption. Mind2Word \cite{mind2word} casts the problem as textual inversion: a Transformer brain encoder followed by a stack of non-linear projections maps fMRI activity into a single pseudo-word token, which is processed by the frozen CLIP text encoder and trained to match the text embedding of the video caption, while auxiliary SoftCLIP losses align the brain representation with both CLIP image and text embeddings, without any fMRI pretraining.\\
Despite their differences, these approaches share a tendency towards increasingly elaborate pipelines that combine several pretrained models, training objectives and auxiliary guidance signals. Our work departs from this trend by returning to the encoding-based perspective of Bayesian decoding \cite{Nishimoto2011, ciferri_bayesian}: instead of enriching the conditioning with additional decoded signals, we keep a single, linearly decoded conditioning and use encoding models to select, among multiple samples of the generator, the one most consistent with the measured brain response.

\section{Methods}
\label{sec:methods}

\subsection{Reranking Approach}
\label{sec:reranking}
\begin{figure}[H]
  \centering
  \includegraphics[width=\linewidth]{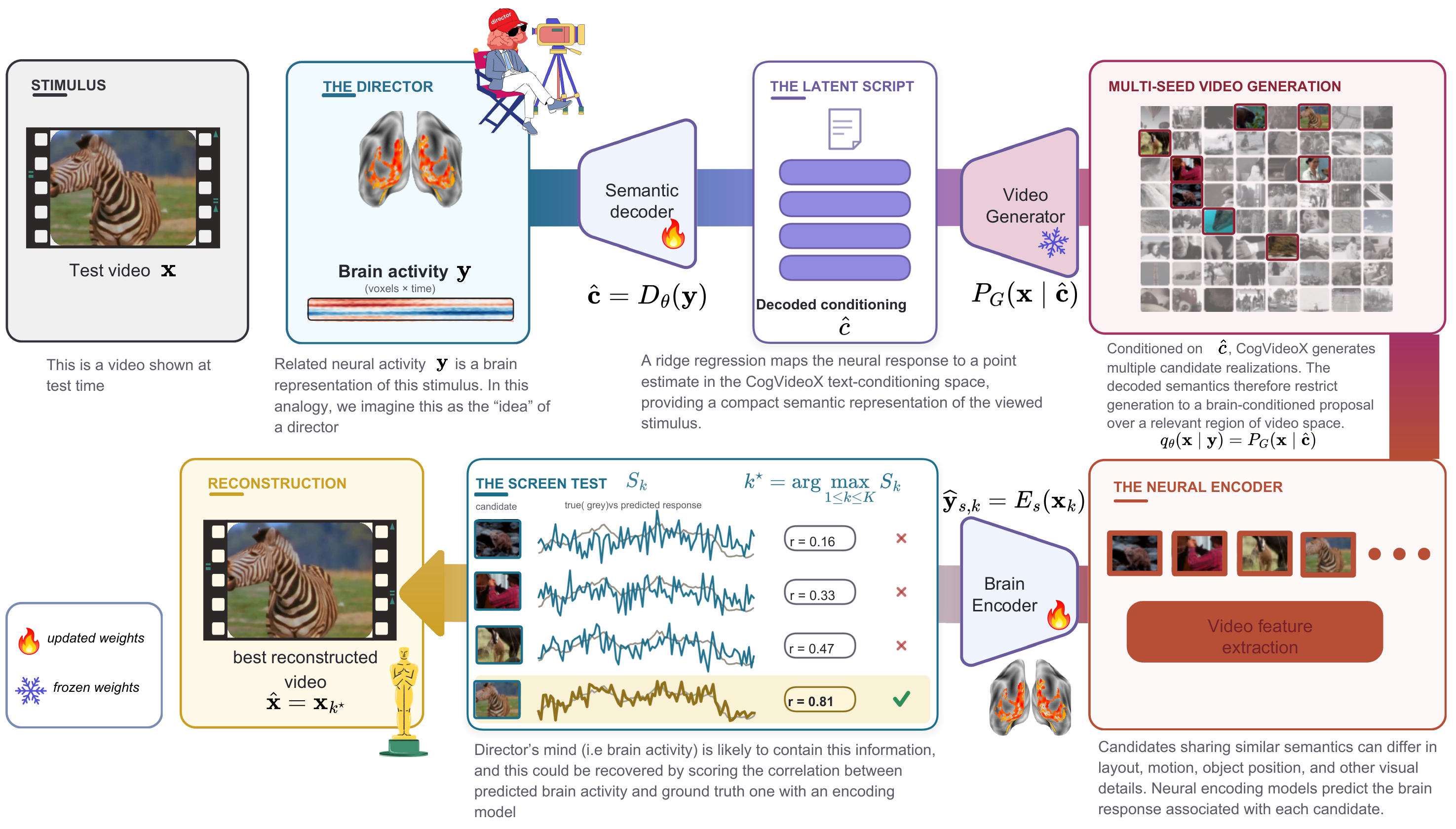}
  \caption{Overview of the proposed decoding framework for both CC2017 and BMD. A linear ridge decoder maps the observed neural response $\mathbf{y}$ to a point estimate $\hat{\mathbf{c}}$ in the CogVideoX text-conditioning space, defining a brain-conditioned proposal $P_G(\mathbf{x}\mid\hat{\mathbf{c}})$. Multiple candidate videos are sampled from this proposal and passed through early, ventral and dorsal neural encoding streams. Their agreement with the observed fMRI response provides neural compatibility scores, and the highest-scoring candidate is selected as the final reconstruction.}
  \label{fig:pipeline}
\end{figure}

Classical Bayesian decoding seeks the posterior $P(\mathbf{x}\mid\mathbf{y})\propto P(\mathbf{y}\mid\mathbf{x})P(\mathbf{x})$, combining neural evidence with a prior over natural videos $\mathbf{x}$ \cite{Nishimoto2011}, but a direct search over such a broad prior is computationally intractable. Following the prior-guided decoding approach of \cite{ciferri_bayesian}, we instead use the observed fMRI response $\mathbf{y}$ to construct a semantically restricted \emph{proposal distribution} over videos, from which a finite set of candidates is drawn and then reranked. The resulting pipeline, illustrated in Figure \ref{fig:pipeline}, consists of three modules.
\paragraph{Semantic Decoder.}
For each training video, the associated caption is encoded in the text-conditioning space of CogVideoX \cite{cogvideo}. For CC2017 we use the captions generated by \cite{NeuroClips} with BLIP-2 \cite{blip2} from the ground-truth videos, whereas for BMD we average the embeddings of the five human-annotated captions provided with each clip in the metadata of \cite{Bold}. A linear ridge decoder $D_\theta$ then maps the selected visual-cortex voxels to each of the target token embeddings: CogVideoX encodes each caption into $226$ T5 tokens of dimension $4096$ (padded to fixed length, without attention mask, so that all $226$ positions are regressed), and for BMD the embeddings of the five captions are averaged token by token before any normalization. One ridge model is fitted per token, with the regularization strength selected by 5-fold cross-validation on the training set for both CC2017 and BMD. \\The predicted embeddings are then renormalized using the mean and standard deviation estimated on the training captions,
\begin{equation}
\hat{\mathbf{c}} = \frac{D_\theta(\mathbf{y})-\boldsymbol{\mu}_{D}}{\boldsymbol{\sigma}_{D}}\odot\boldsymbol{\sigma}_{c}+\boldsymbol{\mu}_{c},
\end{equation}
where $\boldsymbol{\mu}_{D},\boldsymbol{\sigma}_{D}$ are computed on the decoder predictions for the training responses and $\boldsymbol{\mu}_{c},\boldsymbol{\sigma}_{c}$ on the true training embeddings, across training samples and element-wise over the $226\times4096$ token dimensions, yielding a point estimate $\hat{\mathbf{c}}$ in the conditioning space.\\ All model parameters, regularization strengths and normalization statistics are estimated on training data only; test responses enter only as inputs, averaged across repetitions and standardized with training statistics. This estimate defines the brain-conditioned proposal
\begin{equation}
q_\theta(\mathbf{x}\mid\mathbf{y}) := P_G(\mathbf{x}\mid\hat{\mathbf{c}}) = P_G(\mathbf{x}\mid \mathcal{R}(D_\theta(\mathbf{y}))),
\end{equation}
where $P_G$ denotes the conditional distribution of the video generator and $\mathcal{R}$ the renormalization procedure.
\paragraph{Video Generator.}
From the proposal $P_G(\mathbf{x}\mid\hat{\mathbf{c}})$ we generate $K = 100$ candidates by varying the random seed from $42$ to $141$, for every stimulus and subject, using CogVideoX-5b with the default DDIM scheduler \cite{ddim} and bfloat16 precision, with $49$ frames at $720\times480$ pixels, $50$ inference steps and a guidance scale of $6$ (the default) for CC2017 and $3.5$ for BMD. For BMD, whose conditioning is the average of the embeddings of five captions and thus lies outside the distribution of single-sentence embeddings, the default guidance frequently produced static, blurred videos; the lower value was set once by qualitative inspection, applied to all stimuli and subjects, and not tuned on any evaluation metric. Each candidate is exported at $8$ fps. The generated candidates are then temporally downsampled to match the corresponding ground-truth videos, which are represented by $6$ frames uniformly spanning the stimulus: for CC2017, the released ground truth consists of $6$ frames sampled at $3$ fps from each 2-s clip (native $30$ fps) at $260\times259$ pixels; for BMD, we take from the $90$ frames of each 3-s clip ($30$ fps, $268\times268$ pixels) the frames closest to the centers of $6$ equal intervals (frames $8$, $23$, $38$, $53$, $68$ and $83$). The whole $49$-frame candidate is resampled to $6$ frames by trilinear interpolation, without cropping, directly to the resolution of the corresponding ground truth.

\paragraph{Brain Encoder.}
Each candidate is then passed through three neural encoding streams targeting early, ventral and dorsal visual regions; additional details on the regions of interest (ROIs) defined according to scheme A of \cite{cross-subject-video} are provided in Appendix \ref{sec:datasets}. The agreement between predicted and observed neural activity yields a combined compatibility score $S(\mathbf{y},\mathbf{x}_k;\mathbf{X}_{1:K})$, and the final reconstruction is
\begin{equation}
\hat{\mathbf{x}} = \mathbf{x}_{k^\star}, \qquad k^\star = \arg\max_{1\leq k\leq K} S(\mathbf{y},\mathbf{x}_k;\mathbf{X}_{1:K}).
\end{equation}
Each stream combines a pretrained feature extractor with a ridge readout to the fMRI responses of its ROI.\\ The early stream, designed to capture low-level visual details, extracts the \texttt{relu1\_2}, \texttt{relu2\_2}, \texttt{relu3\_3}, \texttt{relu4\_2} and \texttt{relu5\_1} activations of a VGG19-BN network with ImageNet-1K weights \cite{vgg}, applied to $112\times112$ ImageNet-normalized frames following \cite{brain-it}. Each feature map is adaptively pooled to a $7\times7$ grid and aggregated across frames by a learned temporal attention, which uses a separate query for each of the $49$ spatial locations and each feature layer, with the softmax applied only along the temporal dimension.\\ The dorsal stream relies instead on the final hidden states of the V-JEPA~2 ViT-L encoder \cite{vjepa2}, consisting of $2{,}048$ tokens ($8$ temporal $\times\, 16\times16$ spatial) of dimension $1{,}024$, which are aggregated by a learned attention pooler with a single query and four attention heads, using factorized temporal and spatial positional embeddings applied to the keys. Both poolers are trained separately for each subject to predict the fMRI responses of the corresponding ROI through a linear head, using only the training stimuli with an internal validation split for early stopping. For CC2017, the $18$ training movies are divided into six groups of three whole movies, so that neighboring time points never fall on both sides of a split (different movies are acquired in separate runs, hence no temporal gap is needed), and one group is held out; for BMD, a random $12\%$ of the training clips is held out. In both datasets, repetitions of a stimulus are averaged into a single sample. Optimization uses AdamW with a learning rate of $3\times10^{-4}$, a weight decay of $10^{-2}$ and cosine annealing, and the linear heads are discarded after training.\\ Finally, the ventral stream uses the image embedding of OpenCLIP ViT-bigG-14 \cite{clip-vit-big-14} averaged across frames.\\
For each stream, the pooled features are mapped to the fMRI responses of the corresponding ROI using ridge regression, yielding a predicted response for each candidate; the regularization strength is selected on training data only, by cross-validation over the same six movie groups for CC2017 and on a random $15\%$ of the training clips for BMD, and the readout is then refitted on the full training set. Candidates are then scored by computing the Pearson correlation across voxels between each predicted response and the measured fMRI response, averaged across repetitions, followed by $z$-normalization across candidates for each stimulus.
The three stream scores are finally averaged, and the candidate with the highest combined score is selected as the reconstruction.\\ Additional mathematical formulation and implementation details are provided in Appendix \ref{apd:bayesian}.
 
\subsection{Evaluation Metrics}
\label{sec:metrics}
We evaluate the reconstructions both at the frame and the video level, considering local-pixel fidelity and semantic features.\\ At the frame level, pixel-level correspondence is measured with structural similarity index measure (SSIM) \cite{ssim} and peak signal-to-noise ratio (PSNR) \cite{psnr}, while semantic fidelity is assessed through an $N$-way top-$1$ classification test based on an ImageNet classifier \cite{imagenet} (ViT-B/16, \texttt{google/vit-base-patch16-224}; frames resized to $224\times224$): following prior work \cite{mind-video,NeuroClips,NEURONS}, we perform $100$ repeated trials with randomly sampled candidate classes, always including the ground-truth class, and report the classification top-$1$ success rate. The ground-truth classes are not human annotations: they are defined by the classifier itself as the three classes with the highest scores on the original frame. In each trial, $N-1$ distractor classes are drawn at random, without repetition, from the remaining classes. The trial counts as a success if, on the reconstructed frame, the classifier assigns to at least one of the three ground-truth classes a higher probability than to every distractor. Draws are not seeded and thus not shared across methods; scores are averaged over the $100$ trials and over stimuli for each of the $6$ frames, and then over frames. At the video level, semantic fidelity is measured with an analogous $N$-way classification test over the Kinetics-400 classes, using a VideoMAE-based video classifier \cite{videomae} (\texttt{MCG-NJU/videomae-base-finetuned-kinetics}), with pseudo-labels obtained from the original video and scores averaged over trials and stimuli.\\
Finally, CLIP-pcc is the average cosine similarity between the CLIP ViT-L/14 image embeddings of all pairs of distinct frames of the generated video. Because this metric does not compare motion with the ground-truth video, a high score alone does not establish motion fidelity and can also be obtained by nearly static reconstructions.

\section{Results}
\label{sec:results}

\subsection{Comparison with Other Models}
\label{sec:comparison}
Table \ref{tab:comparison} compares the reconstructions obtained with REEL against those of previous fMRI-to-video methods on both datasets, with results averaged across subjects.

\begin{table}[H]
\centering
\small
\renewcommand{\arraystretch}{1.1}

\caption{Comparison between REEL and previous fMRI-to-video methods, grouped by dataset. REEL values are averaged across three subjects for CC2017 and ten for BMD; participant coverage of published baselines follows the original studies, with Brain Netflix evaluated on subject $1$. Bold entries indicate numerical maxima among available comparisons, without implying matched statistical superiority. The rows labelled $K=1$ evaluate the semantic decoder together with the pretrained generator, using one candidate per stimulus without reranking and report the mean across $5$ different measures, as done in Section \ref{effect_size}. On CC2017, NEURONS results were obtained by reproducing its original pipeline, while NeuroClips and MinD-Video results are taken from \cite{NeuroClips,lu2025animatethoughts} and SemVideo results from \cite{semvideo}; on BMD, published values are taken from \cite{brainnetflix,mind2word}. Brain Netflix reports ImageNet-based 2-way/50-way scores of $0.767$/$0.145$ (single-subject training) and $0.816$/$0.165$ (multi-subject training), both evaluated on subject $1$. SemVideo reports a CLIP score of $0.526$ with a VIFI-based gate, so it is not included under our ungated CLIP-pcc definition. A dash denotes an unreported or non-comparable value.}
\label{tab:comparison}
\tablebox{
\resizebox{0.92\textwidth}{!}{%
\begin{tabular}{lccccccc}
\toprule
\textbf{Method} &
\multicolumn{3}{c}{\textbf{Video-based}} &
\multicolumn{4}{c}{\textbf{Frame-based}} \\
\cmidrule(lr){2-4} \cmidrule(lr){5-8}
& \multicolumn{2}{c}{Semantic-level} & ST-level
& \multicolumn{2}{c}{Semantic-level} & \multicolumn{2}{c}{Pixel-level} \\
\cmidrule(lr){2-3} \cmidrule(lr){4-4} \cmidrule(lr){5-6} \cmidrule(lr){7-8}
& 2-way $\uparrow$ & 50-way $\uparrow$ & CLIP-pcc $\uparrow$
& 2-way $\uparrow$ & 50-way $\uparrow$ & SSIM $\uparrow$ & PSNR $\uparrow$ \\
\midrule
\multicolumn{8}{c}{\textbf{CC2017}} \\
\midrule
MinD-Video & $0.8390$ & $0.197$ & $0.408$ & $0.7960$ & $0.174$ & $0.1710$ & $8.662$ \\
NeuroClips & $0.8340$ & $0.2197$ & $0.7377$ & $0.8057$ & $0.2033$ & $0.3903$ & $9.211$ \\
NEURONS & $0.8575$ & $0.2579$ & $\mathbf{0.9367}$ & $0.8083$ & $0.2084$ & $0.3776$ & $\mathbf{9.682}$ \\
SemVideo & $\mathbf{0.865}$ & $\mathbf{0.264}$ & $-$ & $\mathbf{0.838}$ & $\mathbf{0.225}$ & $0.321$ & $9.448$ \\
REEL, $K=1$ (no reranking) & $0.8563$ & $0.2359$ & $0.9180$ & $0.8083$ & $0.1872$ & $0.3951$ & $9.031$ \\
REEL (ours) & $0.8555$ & $0.2434$ & $0.9160$ & $0.8140$ & $0.2049$ & $\mathbf{0.4018}$ & $9.146$ \\
\midrule
\multicolumn{8}{c}{\textbf{BMD}} \\
\midrule
Brain Netflix (single-subject) & $-$ & $-$ & $-$ & $0.767$ & $0.145$ & $0.183$ & $-$ \\
Brain Netflix (multi-subject) & $-$ & $-$ & $-$ & $0.816$ & $0.165$ & $0.190$ & $-$ \\
Mind2Word & $0.842$ & $0.199$ & $-$ & $0.880$ & $0.272$ & $0.26$ & $-$ \\
REEL, $K=1$ (no reranking) & $0.9014$ & $0.3653$ & $0.9111$ & $0.9007$ & $0.3310$ & $0.4060$ & $9.799$ \\
REEL (ours) & $\mathbf{0.9098}$ & $\mathbf{0.3920}$ & $0.9112$ & $\mathbf{0.9097}$ & $\mathbf{0.3535}$ & $\mathbf{0.4116}$ & $9.987$ \\
\bottomrule
\end{tabular}%
}
}
\end{table}
On CC2017, REEL achieves performance comparable to that of other SOTA pipelines. It obtains the highest SSIM among all methods, while on semantic metrics it closely follows the two most recent multi-stage approaches, NEURONS and SemVideo. The latter achieves the best identification accuracies, exceeding REEL by $0.95$--$2.40$ percentage points, whereas the differences with respect to NEURONS are smaller, and REEL even slightly exceeds it in Frame 2-way accuracy. REEL has a CLIP-pcc of $0.9160$, compared with $0.9367$ for our NEURONS reproduction; this describes generated frames feature similarity rather than ground-truth motion fidelity. The published SemVideo score uses a different definition and is not directly comparable. REEL has lower PSNR than NeuroClips, SemVideo and NEURONS. The comparison with the $K=1$ configuration, in which a single candidate is drawn from the brain-conditioned proposal without reranking, shows that the linear semantic decoder alone already accounts for most of this performance, while reranking mainly improves 50-way identification accuracy and structural fidelity, leaving Video 2-way accuracy and CLIP-pcc almost constant; further analysis of the contribution of the reranking method is provided in Section \ref{apd:sampling_scaling} and in Appendix \ref{quant_ablation}.
On BMD, REEL has higher reported identification accuracies and SSIM than the published Mind2Word results shown in Table \ref{tab:comparison}. These are descriptive cross-study comparisons: participant coverage, preprocessing, training regimes, generators, and evaluation implementations are not controlled across methods. Within REEL, increasing from one candidate to $K=100$ with reranking raises mean Video 50-way accuracy from $0.3653$ to $0.3920$ and Frame 50-way accuracy from $0.3310$ to $0.3535$. The strong $K=1$ results show that the combination of semantic decoding and the pretrained generator already accounts for much of the reported performance; the candidate-pool ablation isolates the additional effect of selection.

\subsection{Qualitative Results}
\begin{figure}[H]
\centering
{\includegraphics[width=\linewidth]{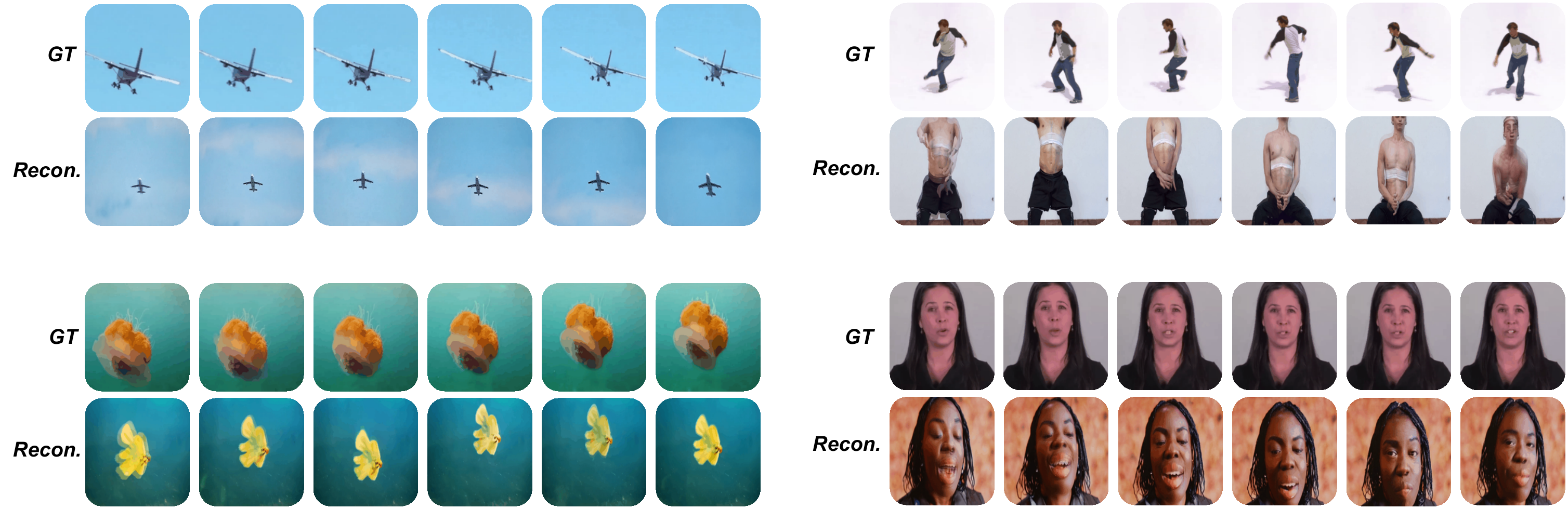}}
  \caption{Manually selected best video reconstructions for CC2017. These frames show a strong correspondence between the original scene shown in the video and our reconstructions, capturing the overall layout, color palette and coarse semantic category.}
   \label{buone_cc}
\end{figure}

\begin{figure}[H]
\centering
{\includegraphics[width=\linewidth]{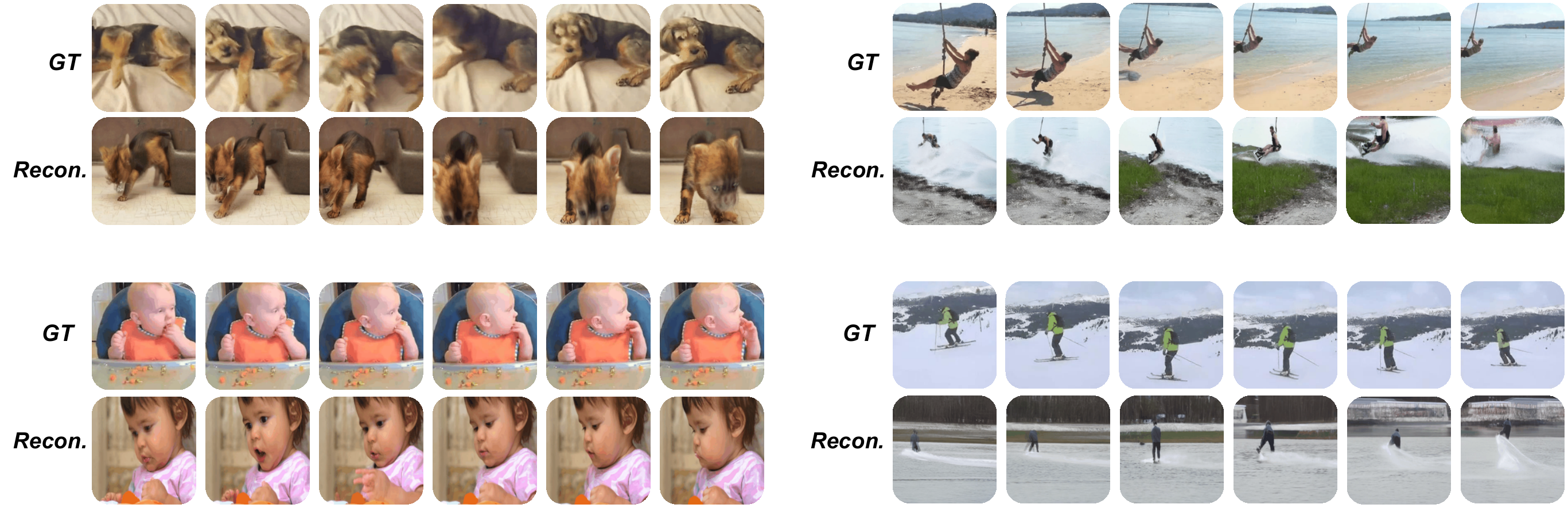}}
  \caption{Manually selected best video reconstructions for BMD. These frames show a strong correspondence between the original scene displayed on the video and our reconstructions, capturing the overall layout, color palette and coarse semantic category.}
   \label{buone_bmd}
\end{figure}

We show manually selected examples of higher reconstruction quality for CC2017 and BMD, in Figures \ref{buone_cc} and \ref{buone_bmd}. These examples are illustrative rather than an estimate of typical performance, since they were selected by visual inspection. Additional examples can be found in Appendix \ref{apd:second} and through the link in the Supplementary Material.

\section{Ablation Studies}
\label{apd:sampling_scaling}
\subsection{Effect of Candidate-Set Size}\label{effect_size}
\begin{figure}[H]
    \centering
    \includegraphics[width=1.0\linewidth]{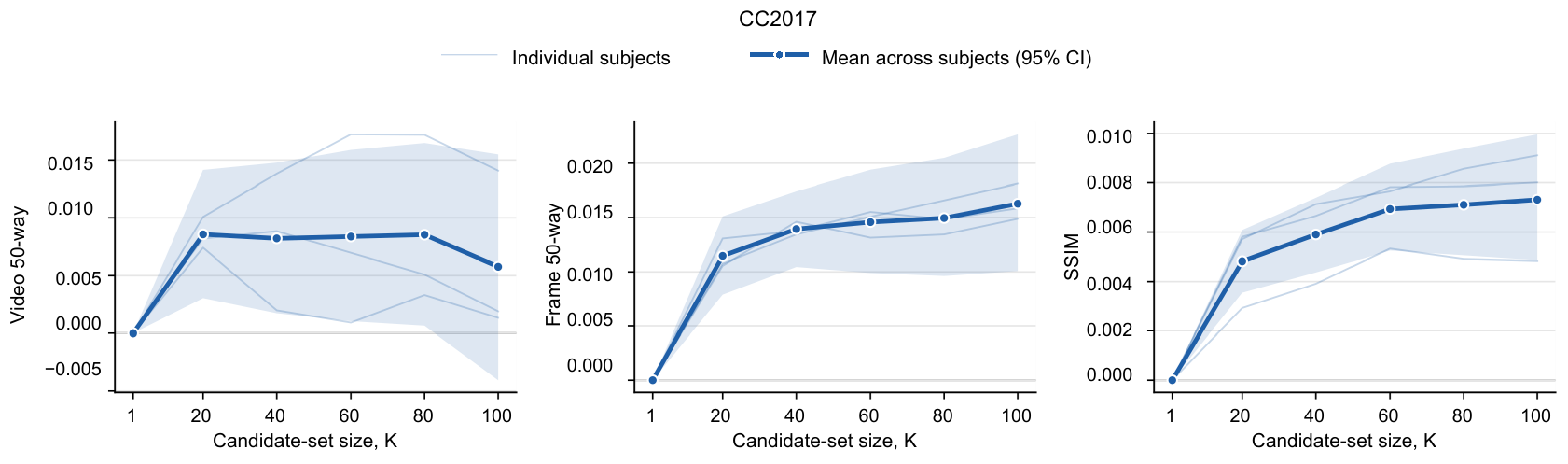}
    \caption{
    \textbf{Effect of candidate-set size on reconstruction performance on CC2017.}
    Paired difference between brain-guided reranking and random selection for Video 50-way accuracy, Frame 50-way accuracy and SSIM, as a function of the number of candidates $K$. For each subject and $K$, the difference is the metric of the reranked candidate minus the random expectation (the mean over the $K$ candidates of each nested subset), computed with per-candidate evaluation metrics and averaged over five subset permutations, which are not treated as independent replicates. Thin lines show the three subjects; the thick line shows their mean, with a 95\% bootstrap confidence interval obtained by resampling blocks of $20$ consecutive test clips within each movie, jointly across subjects. At $K=1$ the two strategies coincide. With three subjects, these results are descriptive.
    }
    \label{fig:ablation-CC2017}
\end{figure}

\begin{figure}[H]
    \centering
    \includegraphics[width=1.0\linewidth]{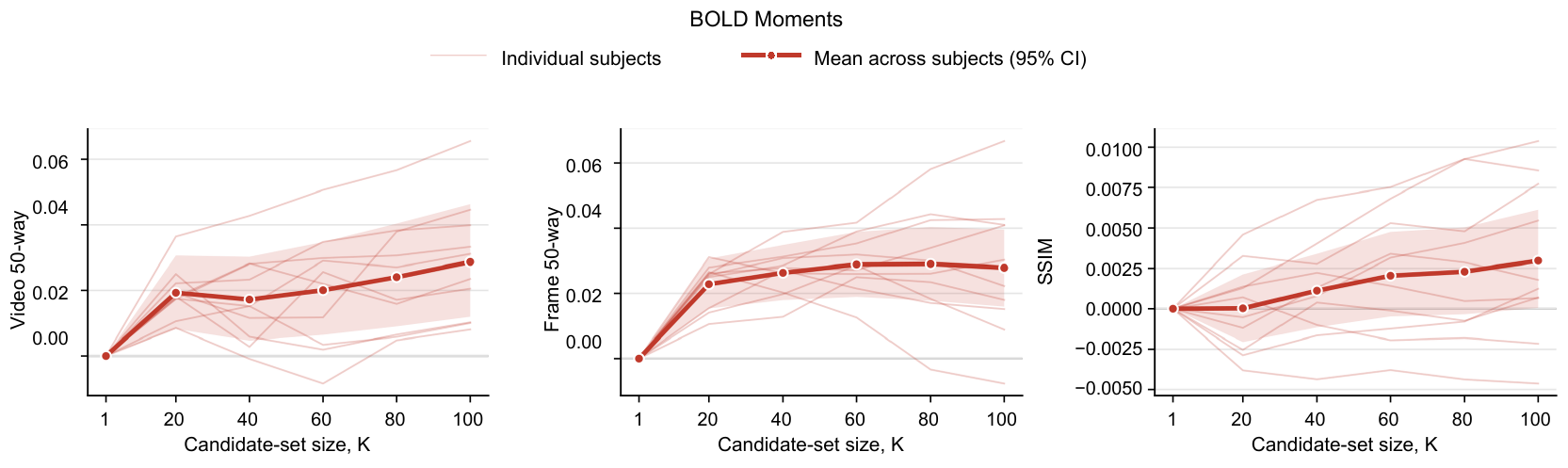}
    \caption{
    \textbf{Effect of candidate-set size on reconstruction performance on BMD.}
    Paired difference between brain-guided reranking and random selection for Video 50-way accuracy, Frame 50-way accuracy and SSIM, as a function of the number of candidates $K$. For each subject and $K$, the difference is the metric of the reranked candidate minus the random expectation (the mean over the $K$ candidates of each nested subset), computed with per-candidate evaluation metrics and averaged over five subset permutations, which are not treated as independent replicates. Thin lines show the ten subjects; the thick line shows their mean, with a $95\%$ bootstrap confidence interval obtained by resampling test clips jointly across subjects. At $K=1$ the two strategies coincide.
    }
    \label{fig:ablation-bmd}
\end{figure}

To disentangle the contribution of brain-guided selection from that of the semantic proposal itself, we examined how reconstruction quality changes as the number of generated candidates $K$ increases, comparing the candidate selected by the encoding models with one drawn at random from the same pool.
To isolate the effect of pool size from that of score calibration, the standardization statistics are computed once per stimulus on the full pool of $100$ candidates and kept fixed for all subsets, so that each candidate receives the same score regardless of the subset it belongs to. The curves therefore measure how much the same selector gains when choosing from a larger pool, rather than the performance of the full procedure run with only $K$ candidates. 
For a fixed proposal, uniform random selection has the same expected quality for every $K$. Comparing reranking with random selection from the same pool therefore measures the effect of the selection rule, although the observed differences retain finite-sample variability. Figures \ref{fig:ablation-CC2017} and \ref{fig:ablation-bmd} report, for Video 50-way accuracy, Frame 50-way accuracy and SSIM, the paired difference between the reranked candidate and the random expectation, i.e., the mean over the $K$ candidates of the same subset, computed on the same stimuli and candidate subsets for each subject; absolute values from the official evaluation are reported in Appendix \ref{quant_ablation}. 

On both datasets, the reconstructions selected by the encoding models are more faithful to the ground truth than random selection from the same pool at nearly $K>1$, at the semantic as well as at the structural level. On CC2017, increasing the candidate pool from $K=1$ to $K=100$ raises Video 50-way accuracy by $0.75$ percentage points ($3.2$\% relative), Frame 50-way accuracy by $1.77$ percentage points ($9.5$\% relative), and SSIM by $0.0067$ ($1.7\%$ relative). On BMD, the corresponding accuracy gains are $2.67$ and $2.25$ percentage points ($7.3\%$ and $6.8\%$ relative). The improvement in SSIM is smaller on BMD, where the reranked and randomly selected candidates differ by about $0.003$ at $K=100$. The paired differences show that the gain is not strictly monotonic in every case: on CC2017, the Video 50-way difference is approximately constant from $K=20$ to $K=80$ and smaller at $K=100$, where its confidence interval includes zero, while Frame 50-way accuracy and SSIM increase with $K$ for all three subjects; on BMD, the Video 50-way increases overall from $K= 40$, whereas the Frame 50-way difference plateaus from $K=60$. Early gains also depend on the metric: at $K = 20$, CC2017 Video 50-way accuracy already exceeds its $K=1$ to $K=100$ improvement, whereas BMD Video 50-way accuracy has attained about $36\%$ of it and continues to improve at larger pool sizes. This behavior could be explained with the fact that the reranking  is not guaranteed to translate monotonically into external evaluation metrics.\\ To test whether the gain relies on stimulus-specific information in the measured response, we repeated the selection keeping each stimulus's candidates and conditioning fixed but scoring them against the response to a different test stimulus (for CC2017, always from a different movie). With shuffled responses, semantic accuracy falls below random selection, whereas with the reranked response it exceeds both, as reported in Appendix \ref{apd:shuffle}.

\subsection{Contribution of Brain Encoding Streams}
\label{sec:streams}

We then assessed the contribution of each encoding stream through an ablation in which the candidates are reranked using either a single stream or a pair of streams, and the resulting reconstructions are compared with those of the full model, which combines all three streams. Since each configuration selects among the same $K=100$ candidates, any difference in performance reflects only the choice of the streams used for scoring. Table \ref{tab:ablation-encoder} reports, for each configuration, the relative change with respect to the full model, whose absolute values are given in the bottom row. These comparisons were not prespecified and the tests are not corrected for multiple comparisons, so we treat them as exploratory and also report how consistently each effect appears across subjects.
 
\begin{table}[H]
\centering
\small
\renewcommand{\arraystretch}{1.1}

\caption{Contribution of each encoding stream and of each pair of streams on CC2017 (averaged over $3$ subjects) and BMD (averaged over $10$ subjects). Values report the relative change (\%) with respect to the full model, which combines all three streams (bottom row, absolute values); a negative value indicates that removing the remaining stream(s) degrades reconstruction quality. All configurations select among the same $K=100$ candidates. Markers denote uncorrected paired $t$-tests across subjects ($\mathrm{df}=2$ for CC2017, $\mathrm{df}=9$ for BMD) and should be read as exploratory: \textsuperscript{*}~$p<0.05$, \textsuperscript{**}~$p<0.01$; no marker denotes $p \geq 0.05$.}
\label{tab:ablation-encoder}
\tablebox{
\resizebox{0.92\textwidth}{!}{%
\begin{tabular}{lcccccccc}
\toprule
\textbf{Encoder} &
\multicolumn{4}{c}{\textbf{CC2017}} &
\multicolumn{4}{c}{\textbf{BMD}} \\
\cmidrule(lr){2-5} \cmidrule(lr){6-9}
& Video 50-way & Frame 50-way & SSIM & PSNR
& Video 50-way & Frame 50-way & SSIM & PSNR \\
\midrule
Early & $-2.6\%$ & $-6.4\%$\rlap{\textsuperscript{**}} & $+1.4\%$\rlap{\textsuperscript{*}} & $+0.2\%$ & $-7.4\%$\rlap{\textsuperscript{**}} & $-2.0\%$ & $+0.4\%$ & $+0.0\%$ \\
Ventral & $+4.0\%$ & $-0.9\%$ & $-2.4\%$\rlap{\textsuperscript{*}} & $-1.2\%$\rlap{\textsuperscript{**}} & $-3.7\%$ & $+0.6\%$ & $-1.7\%$\rlap{\textsuperscript{*}} & $-1.3\%$ \\
Dorsal & $-1.0\%$ & $-3.5\%$ & $-1.1\%$ & $-0.4\%$ & $-4.3\%$ & $-3.5\%$ & $-1.2\%$\rlap{\textsuperscript{**}} & $-0.5\%$ \\
\midrule
Early + Ventral & $+1.5\%$ & $-2.2\%$ & $+0.3\%$ & $-0.2\%$ & $-2.4\%$ & $+0.2\%$ & $-0.2\%$ & $-0.5\%$ \\
Early + Dorsal & $-3.0\%$ & $-2.2\%$ & $+0.5\%$ & $+0.4\%$ & $-2.1\%$ & $-0.3\%$ & $-0.3\%$ & $-0.6\%$\rlap{\textsuperscript{*}} \\
Ventral + Dorsal & $+3.8\%$ & $+0.1\%$ & $-1.6\%$\rlap{\textsuperscript{**}} & $-0.2\%$ & $-1.3\%$ & $+0.3\%$ & $-1.7\%$\rlap{\textsuperscript{**}} & $-1.0\%$\rlap{\textsuperscript{*}} \\
\midrule
All streams & $0.2434$ & $0.2049$ & $0.4018$ & $9.146$ & $0.3920$ & $0.3535$ & $0.4116$ & $9.987$ \\
\bottomrule
\end{tabular}%
}
}
\end{table}

On BMD, the full model has a higher Video 50-way accuracy than any single-stream configuration: reranking with any individual stream reduces it by $3.7\%$ to $7.4\%$, suggesting that semantic identification benefits from integrating information across visual regions, although at the subject level this reduction is consistent only for the early stream (nine of ten subjects). On CC2017 the pattern is less consistent: the early and dorsal streams reduce Video 50-way accuracy by $2.6\%$ and $1.0\%$, respectively, whereas the ventral stream alone increases it by $4.0\%$, although none of these differences is significant. The early stream shows the clearest dissociation between the two levels of description. Used alone, it causes the largest semantic losses, namely in Frame 50-way accuracy on CC2017 ($-6.4\%$, in all three subjects) and in Video 50-way accuracy on BMD ($-7.4\%$, in nine of ten subjects), while preserving, or even slightly improving, structural fidelity, with SSIM changes of $+1.4\%$ on CC2017 (positive in all three subjects) and $+0.4\%$ on BMD (not significant). The ventral and dorsal streams display the opposite behavior, as both reduce SSIM on both datasets when used alone, with the same sign in all three CC2017 subjects and in nine of ten BMD subjects. Between the two, the ventral stream best preserves frame-level semantics, remaining within $2\%$ of the full model in Frame 50-way accuracy on both datasets, and on CC2017 it even exceeds the full model in Video 50-way accuracy ($+4.0\%$, not significant), while the dorsal stream loses only $1.0\%$. These patterns are broadly consistent with prior descriptions of visual cortical organization \cite{early1,early2,dorsal,ventral1,ventral2}. However, each stream changes the target ROI, feature extractor, and pooling procedure together, so the ablation does not isolate a cortical region's contribution or establish motion-specific information in the dorsal stream.\\
The pairwise configurations confirm this picture. Pairs that include the early stream remain close to the full model on structural metrics, whereas removing it, as in the Ventral + Dorsal configuration, is the only pairwise ablation that reduces SSIM on both datasets ($-1.6\%$ on CC2017, in all three subjects, and $-1.7\%$ on BMD, in nine of ten subjects), together with a decrease in PSNR ($-0.2\%$ on CC2017, not significant, and $-1.0\%$ on BMD. On BMD, pairs generally lose less than single streams on semantic metrics, with Video and Frame 50-way accuracy remaining within $2.5\%$ of the full model in all cases, which suggests that the three streams provide complementary rather than redundant information, although subject-level changes in 50-way accuracy are of mixed sign. On CC2017, pairwise changes in Video 50-way accuracy range from $-3.0\%$ (Early + Dorsal) to $+3.8\%$ (Ventral + Dorsal), none of them significant. The combination of semantic and structural information nonetheless involves a mild trade-off: on CC2017, the early stream alone achieves a slightly higher SSIM than the full model ($+1.4\%$, in all three subjects), at the cost of the largest drop in Frame 50-way accuracy ($-6.4\%$), showing a small selection trade-off under the current scoring rule. On BMD, this trade-off remains small, since the full model achieves the highest Video 50-way accuracy in every comparison and lies within $1\%$ of the best configuration on every other metric. On CC2017, several configurations exceed the full model in Video 50-way accuracy (up to $+4.0\%$), but none of the improvements is significant; we therefore retain the uniform combination of the three streams as a simple default rather than an optimized choice. Given the small number of subjects in CC2017, these stream-specific effects should nonetheless be regarded as indicative rather than conclusive for this dataset.

\section{Discussion}
\label{sec:discussion}
 
We introduced REEL, a simple framework for video decoding from fMRI that combines semantic decoding, generative proposal sampling and neural reranking, and that achieves performance comparable to SOTA models in literature. The key insight behind it is that a linear mapping from brain activity to the text-conditioning space of a pretrained video generator is sufficient to restrict generation to a semantically relevant region of video space, from which multiple candidate realizations can be sampled and then discriminated by encoding models that compare their predicted neural responses with the observed fMRI activity. This separation between \emph{where to search} and \emph{which candidate to select} offers a simple and interpretable alternative to increasingly complex multi-conditioning decoding pipelines, and makes the role of each component directly measurable.\\
Although motivated by classical Bayesian decoding, which combines an intractable generative prior with an encoding model, our framework does not explicitly evaluate the exact posterior $P(\mathbf{x}\mid\mathbf{y})$. The decoded semantic representation $\hat{\mathbf{c}}=\mathcal{R}(D_\theta(\mathbf{y}))$ instead defines the brain-conditioned proposal $q_\theta(\mathbf{x}\mid\mathbf{y}) = P_G(\mathbf{x}\mid\hat{\mathbf{c}})$, from which a tractable set of candidates is sampled. Further, the three encoding models, each specialized in predicting the activity of a different visual region, provide relative compatibility scores over this finite set. Semantic decoding constructs the proposal, whereas encoding-based reranking uses the measured response to select among its samples. The ablations in Section \ref{apd:sampling_scaling} and Appendix \ref{quant_ablation} show that this selection can improve reconstruction metrics.\\
It is worth noting that, even with a single candidate video, that is, without any contribution from the brain encoder, the results are already close to those of other SOTA methods on CC2017 and even above them on BMD. This difference between the two datasets, which is also preserved with the full pipeline, is likely due to their distinct designs. In CC2017 the clips are segments cut from longer continuous video sequences, so that a single fMRI sample may span scene transitions and partially overlapping content, whereas in BMD each clip depicts a single self-contained event. Moreover, the captions used to train the semantic decoder were generated automatically with BLIP-2 by \cite{NeuroClips} for CC2017, while for BMD five captions were independently written by human annotators for each clip; averaging their embeddings may provide less noisy semantic targets. These explanations remain hypotheses because acquisition design, preprocessing, caption supervision, and stimulus content differ together between the datasets.\\
The NEURONS candidate-pool experiment in Appendix \ref{neurons_experiment} shows that reranking can also increase mean video-level semantic accuracy in another pipeline. One possible explanation for the smaller gains is lower candidate diversity under stronger guidance. This remains an
hypothesis because the pipelines also differ in generator, conditioning, and candidate construction.
 
\section{Limitations and Conclusions}
\label{sec:conclusions}
 
A major limitation of the proposed approach is the computational cost of generating and evaluating multiple candidate videos for each neural sample, which amounts to about $2.5$ min per video-candidate on a single H100 GPU, and may restrict its accessibility in resource-constrained settings. Moreover, reranking can only select among the candidates produced by the generator, so that its performance is ultimately bounded by the quality of the proposal: when a stimulus falls outside the semantic range that the decoder and the generator can represent, no candidate in the pool may be close to the ground truth. The compatibility score is also defined relative to the candidate set rather than as a calibrated likelihood, which prevents a direct probabilistic interpretation of the selected reconstruction.\\
The reported test reconstructions use fMRI responses averaged across ten presentations of each stimulus; performance from a single presentation or in an online setting has not been established. The current framework is also subject-specific, motivating future work on shared decoders and encoding models capable of generalizing across individuals. Future extensions could also represent the uncertainty of the decoded semantic conditioning rather than relying on a single point estimate, allowing the proposal distribution to account for uncertainty in both semantic decoding and video generation.\\
A further important consideration concerns the broader implications of these developments. As generative models continue to advance, video brain decoding could provide increasingly powerful tools for investigating dynamic visual representations and advancing future brain--computer interfaces. Our results suggest that, beyond increasingly sophisticated architectures, relatively simple pipelines that explicitly leverage encoding models can offer a competitive and interpretable approach to this goal. This perspective may help guide the development of future decoding systems, while underscoring the need for appropriate safeguards to protect neural-data privacy and ensure the responsible use of increasingly capable decoding technologies.

\subsection*{AI use statement}
In this work, we used generative AI tools for polishing text in English.
We have not used generative AI tools for generating concepts and procedure ideas. We have reviewed all AI-assisted work. Part of the code was LLM-generated and was verified and tested for correctness
by the authors. We take responsibility for the final content of this work,
including text, claims or artifacts produced with the aid of generative AI.

\section*{Supplementary Material}\label{videoextra}
Additional qualitative examples of reconstructed videos are reported at the following link: \url{https://github.com/fabriziospera/video_decoding_fMRI_reconstructions}. All videos were selected through visual inspection by the authors.

\bibliographystyle{ieeetr}
\bibliography{bib}

\appendix

\section{Datasets and Preprocessing}
\label{sec:datasets}
The CC2017 dataset contains $18$ training videos and $5$ held-out test videos, each $8$ minutes long, which were presented to three participants $2$ and $10$ times, respectively; responses were averaged across repetitions to obtain a more reliable estimate of the neural response. Data were acquired on a $3$-T MRI scanner with a repetition time of $2$-s, yielding $4320$ training and $1200$ test fMRI-video pairs ($240$ volumes per video, each paired with a $2$-s clip).\\ We use the fMRI data released with the dataset, already preprocessed according to the minimal pipeline adopted by NeuroClips \cite{NeuroClips}, which includes artifact removal, six-degree-of-freedom motion correction, alignment to the MNI standard space and projection into cortical surfaces registered to a common template. The released data are restricted to voxels responsive to the visual stimuli, selected by quantifying the consistency of voxel-wise responses across the two presentations of each training video: for each voxel and each of the 18 training videos, the Pearson correlation between the two repetitions was computed across time points and Fisher $z$-transformed, yielding $18$ $z$ values per voxel. These values were tested against zero with a voxel-wise two-sided one-sample $t$-test ($\mathrm{df}=17$), and voxels surviving Bonferroni correction ($P<0.05$) were retained as stimulus-responsive, resulting in $13447$, $14828$ and $9114$ voxels in the visual cortex of the three subjects, respectively. Finally, to account for the delay of the hemodynamic response, BOLD measurements were shifted by $4$-s with respect to the corresponding video stimuli, following previous fMRI-to-video decoding studies \cite{delay_bold1,delay_bold2}; this alignment is already included in the released data, and no further shift was applied. Since each run continues after the end of the movie, the delayed responses to the last clips remain within the same run: all $240$ pairs of each run are retained and no pair spans two runs.\\
The BMD \cite{Bold} contains $1102$ distinct $3$-s video clips drawn from the Memento10k video memorability dataset \cite{memento10k}, presented to $10$ participants in an event-related design, with $1000$ training clips repeated three times each and $102$ test clips repeated ten times each. Data were acquired on a $3$-T Siemens Prisma scanner (TR = $1.75$ s, $2.5$ mm voxels) and, as for CC2017, responses were averaged across repetitions, yielding $1000$ training and $102$ test fMRI-video pairs per participant. We use the preprocessed data released by the authors (OpenNeuro, version B derivatives), obtained with fMRIPrep and projected into the fsLR32k cortical surface. Since single-trial responses were estimated by the authors with GLMsingle, which fits a voxel-specific hemodynamic response function together with data-driven denoising and ridge regularization, no temporal shift was applied; the resulting betas were $z$-scored within each session using the statistics of the training trials, applied to both training and test trials. Voxel selection follows the CC2017 procedure, adapted to the event-related design, and is performed on the individual repetitions, before averaging. Because repeated presentations of a clip are not temporally aligned, response consistency was computed across stimuli rather than across time: the $1000$ training clips were randomly partitioned into $18$ blocks of $55$--$56$ clips, mirroring the $18$ training videos of CC2017 and within each block we computed the correlation between repetitions across clips, which was then Fisher $z$-transformed and averaged over the three pairs of repetitions, yielding $18$ $z$ values per voxel. As for CC2017, these values were tested against zero with a voxel-wise two-sided one-sample $t$-test ($\mathrm{df}=17$), and voxels with a positive $t$ surviving Bonferroni correction were retained, resulting in $1538$ to $5161$ selected voxels per participant. The number of blocks was fixed a priori. Test stimuli were never used for voxel selection in either dataset. Table \ref{tab:data-preprocessing} summarizes the splits and preprocessing of the two datasets.
\begin{table}[H]
\centering
\small
\renewcommand{\arraystretch}{1.1}

\caption{Summary of data splits and preprocessing for the two datasets.}
\label{tab:data-preprocessing}
\tablebox{
\begin{tabular}{p{0.22\linewidth}p{0.35\linewidth}p{0.35\linewidth}}
\toprule
 & \textbf{CC2017} & \textbf{BMD} \\
\midrule
Stimuli & $18$ training $+$ $5$ test movies ($8$ min each) & $1000$ training $+$ $102$ test clips ($3$ s each) \\
Design / TR & continuous movies, TR = $2$ s & event-related, randomized order, TR = $1.75$ s \\
Repetitions (train/test) & $2$/$10$, averaged & $3$ / $10$, averaged \\
Pairs (train/test) & $4320$/$1200$ ($240$ volumes per movie, $2$-s clips) & $1000$ / $102$ \\
Released data & NeuroClips minimal pipeline, cortical surface & fMRIPrep, fsLR32k surface (version B) \\
Response unit & fMRI volume (time point) & single-trial GLMsingle beta \\
Hemodynamic delay & $4$-s shift, already in the released data & modelled by GLMsingle, no shift \\
Normalization & as released & $z$-scored per session with training statistics \\
Voxel selection unit & $18$ training movies, correlation across time & $18$ random blocks of training clips, correlation across clips \\
Selection test & two-sided one-sample $t$-test, df = $17$, Bonferroni $P<0.05$ & same, positive $t$ only \\
Selected voxels & $9114$--$14828$ (released) & $1538$--$5161$ \\
\bottomrule
\end{tabular}
}
\end{table}

\begin{table}[H]
\centering
\small
\renewcommand{\arraystretch}{1.1}

\caption{Assignment of HCP-MMP1 areas \cite{glasser_multi-modal_2016} to the three visual streams. All streams are bilateral.}
\label{tab:roi-areas}
\tablebox{
\begin{tabular}{lp{0.68\linewidth}}
\toprule
\textbf{Stream} & \textbf{Areas} \\
\midrule
Early   & V1, V2, V3, V4 \\
Dorsal  & V3A, V3B, V6, V6A, V7, IPS1, LO1, LO2, LO3, FST, MT, MST, V3CD, V4t, PH, IP0 \\
Ventral & FFC, PIT, V8, VMV1, VMV2, VMV3, VVC, PHA1, PHA2, PHA3, TE2p \\
\bottomrule
\end{tabular}
}
\end{table}

\section{Bayesian Motivation and Brain-Guided Proposal Search}
\label{apd:bayesian}

\begin{figure}[H]
{\includegraphics[width=\linewidth]{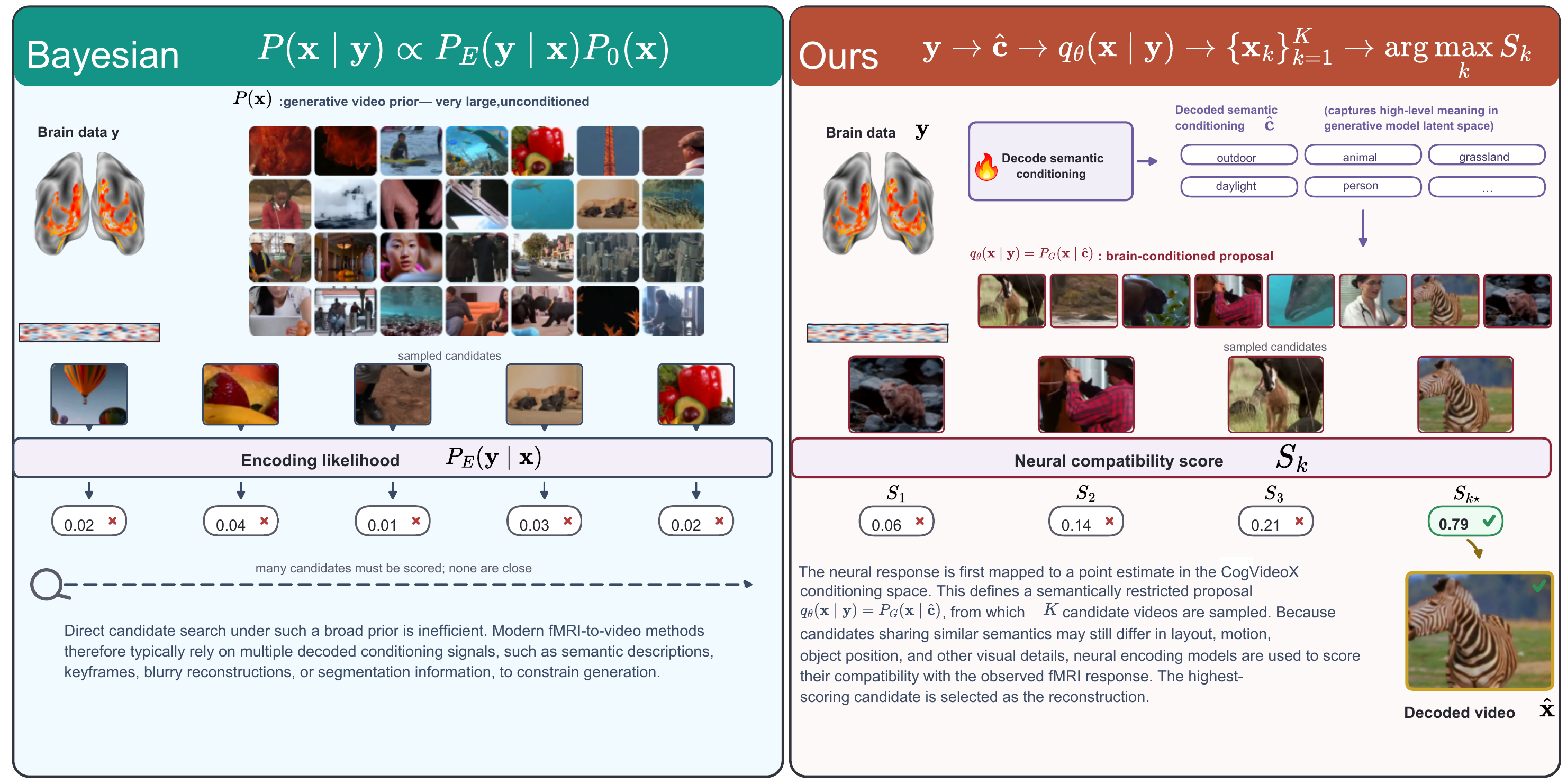}}
   \caption{Bayesian motivation for our proposal-and-rerank strategy.
Classical Bayesian decoding combines a broad prior over natural videos with
neural evidence. Direct candidate search by unconditional sampling from this
prior is impractical. Our approach instead uses the observed brain response to
construct a semantically restricted proposal distribution, samples a tractable
set of candidate videos, and reranks them using neural encoding models.}
   \label{fig:bayesian_comparison}
\end{figure}

\subsection{Bayesian motivation}

Let $\mathbf{x}$ denote a visual stimulus and $\mathbf{y}$ the corresponding
measured neural response. Classical Bayesian decoding targets the posterior
\begin{equation}
P(\mathbf{x}\mid\mathbf{y})
=
\frac{
P_E(\mathbf{y}\mid\mathbf{x})P_0(\mathbf{x})
}{
P(\mathbf{y})
}
\propto
P_E(\mathbf{y}\mid\mathbf{x})P_0(\mathbf{x}),
\label{eq:bayes_classical}
\end{equation}
where $P_0(\mathbf{x})$ is a prior over natural videos and
$P_E(\mathbf{y}\mid\mathbf{x})$ is an encoding likelihood describing how a
stimulus gives rise to neural activity.

Equation \ref{eq:bayes_classical} provides useful conceptual guidance, but
direct posterior search is impractical for modern video spaces. A pretrained
video generator places probability mass over an extremely broad range of
objects, scenes, viewpoints, motions, and visual styles. Unconditional
sampling from this space would therefore require a prohibitively large
candidate set before obtaining videos that are even semantically compatible
with a particular neural observation.

Introducing a semantic variable $\mathbf{c}$ does not in general permit the
identity
$P_0(\mathbf{x}) =
P(\mathbf{x}\mid\mathbf{c})P(\mathbf{c}\mid\mathbf{y})$.
If $\mathbf{c}$ were a latent variable of the generative prior, the
corresponding marginal prior would instead be
\begin{equation}
P_0(\mathbf{x})
=
\int
P_G(\mathbf{x}\mid\mathbf{c})P_0(\mathbf{c})\,d\mathbf{c}.
\label{eq:latent_prior}
\end{equation}
Our method takes a different, observation-dependent route: the neural signal
is used to construct a \emph{proposal distribution} that concentrates
generation on a semantically relevant region of the video space.

\subsection{Semantic decoding and the brain-conditioned proposal}

For each training example, let
$\mathbf{y}_i\in\mathbb{R}^{V}$ denote the selected fMRI voxel response and
$\mathbf{c}_i\in\mathbb{R}^{D}$ the corresponding target conditioning
representation in the CogVideoX text-conditioning space.

Writing the training matrices as
\begin{equation}
\mathbf{Y}
=
[\mathbf{y}_1^\top,\ldots,\mathbf{y}_N^\top]^\top
\in\mathbb{R}^{N\times V},
\qquad
\mathbf{C}
=
[\mathbf{c}_1^\top,\ldots,\mathbf{c}_N^\top]^\top
\in\mathbb{R}^{N\times D},
\end{equation}
we fit a multi-output ridge decoder
\begin{equation}
(\widehat{\mathbf{B}},\widehat{\mathbf{b}})
=
\arg\min_{\mathbf{B},\mathbf{b}}
\left\|
\mathbf{C}
-
\mathbf{Y}\mathbf{B}
-
\mathbf{1}\mathbf{b}^{\top}
\right\|_F^2
+
\mathbf{\lambda_t}\|\mathbf{B}\|_F^2,
\label{eq:semantic_ridge}
\end{equation}
where $\mathbf{\lambda_t},\mathbf{B}\in\mathbb{R}^{V\times D}$ and
$\mathbf{b}\in\mathbb{R}^{D}$. This applies for each token.

For a new neural response $\mathbf{y}$, the decoder produces
\begin{equation}
\mathbf{c}_{\mathrm{raw}}
=
D_\theta(\mathbf{y})
=
\widehat{\mathbf{B}}^\top\mathbf{y}
+
\widehat{\mathbf{b}},
\end{equation}
followed by the conditioning-space renormalization,
which we denote by $\mathcal{R}$:
\begin{equation}
\hat{\mathbf{c}}
=
\mathcal{R}\!\left(D_\theta(\mathbf{y})\right).
\label{eq:c_hat}
\end{equation}

The point estimate $\hat{\mathbf{c}}$ is kept fixed during candidate
generation. It induces the conditional proposal
\begin{equation}
q_\theta(\mathbf{x}\mid\mathbf{y})
:=
P_G(\mathbf{x}\mid\hat{\mathbf{c}})
=
P_G\!\left(
\mathbf{x}
\mid
\mathcal{R}(D_\theta(\mathbf{y}))
\right),
\label{eq:proposal}
\end{equation}
where $P_G$ denotes the conditional video distribution represented by
CogVideoX.

An equivalent probabilistic notation makes the role of the point decoder
explicit. Define
\begin{equation}
q_\theta(\mathbf{c}\mid\mathbf{y})
=
\delta\!\left(
\mathbf{c}-\hat{\mathbf{c}}
\right),
\end{equation}
where $\delta$ is a Dirac point mass. Then
\begin{equation}
q_\theta(\mathbf{x}\mid\mathbf{y})
=
\int
P_G(\mathbf{x}\mid\mathbf{c})
q_\theta(\mathbf{c}\mid\mathbf{y})
\,d\mathbf{c}
=
P_G(\mathbf{x}\mid\hat{\mathbf{c}}).
\label{eq:proposal_integral}
\end{equation}

\paragraph{Proposition 1 (validity of the semantic proposal).}
If $P_G(\mathbf{x}\mid\mathbf{c})$ is a normalized conditional distribution
over $\mathbf{x}$ for every $\mathbf{c}$, and
$q_\theta(\mathbf{c}\mid\mathbf{y})$ is a normalized conditional distribution
over $\mathbf{c}$, then
\begin{equation}
q_\theta(\mathbf{x}\mid\mathbf{y})
=
\int
P_G(\mathbf{x}\mid\mathbf{c})
q_\theta(\mathbf{c}\mid\mathbf{y})
\,d\mathbf{c}
\end{equation}
is itself a normalized conditional distribution over videos.

\paragraph{Proof.}
Non-negativity follows immediately from
$P_G(\mathbf{x}\mid\mathbf{c})\geq 0$ and
$q_\theta(\mathbf{c}\mid\mathbf{y})\geq 0$.
Moreover, since the integrand is non-negative, we may exchange the order
of integration:
\begin{align}
\int q_\theta(\mathbf{x}\mid\mathbf{y})\,d\mathbf{x}
&=
\int
\int
P_G(\mathbf{x}\mid\mathbf{c})
q_\theta(\mathbf{c}\mid\mathbf{y})
\,d\mathbf{c}\,d\mathbf{x}
\\
&=
\int
q_\theta(\mathbf{c}\mid\mathbf{y})
\left[
\int
P_G(\mathbf{x}\mid\mathbf{c})\,d\mathbf{x}
\right]
d\mathbf{c}
\\
&=
\int
q_\theta(\mathbf{c}\mid\mathbf{y})
\,d\mathbf{c}
\\
&=1,
\end{align}
where the reduction in the third line follows from the normalization of
$P_G(\mathbf{x}\mid\mathbf{c})$, and the final equality follows from the
normalization of $q_\theta(\mathbf{c}\mid\mathbf{y})$.
Hence $q_\theta(\mathbf{x}\mid\mathbf{y})$ is a valid conditional probability
distribution over $\mathbf{x}$. \hfill$\square$

For the deterministic ridge decoder,
$q_\theta(\mathbf{c}\mid\mathbf{y})
=\delta(\mathbf{c}-\hat{\mathbf{c}})$, so the proposal reduces directly to
\begin{equation}
q_\theta(\mathbf{x}\mid\mathbf{y})
=
P_G(\mathbf{x}\mid\hat{\mathbf{c}}),
\end{equation}
as in Equation \ref{eq:proposal}.

We then draw
\begin{equation}
\mathbf{x}_1,\ldots,\mathbf{x}_K
\overset{\mathrm{i.i.d.}}{\sim}
q_\theta(\mathbf{x}\mid\mathbf{y}),
\qquad K=100.
\label{eq:candidate_sampling}
\end{equation}
Thus, semantic decoding does not itself specify a unique reconstruction.
Instead, it determines \emph{where in the generative video space to search}.

\subsection{Neural encoding and candidate compatibility}

Semantic conditioning substantially restricts the search space, but it does
not determine the precise visual realization of a stimulus. Videos generated
from the same $\hat{\mathbf{c}}$ may share their broad semantic content while
differing in other fine details, such as structural information, colors and illumination.

We therefore return to the measured neural response to discriminate among the
sampled hypotheses. Let
\begin{equation}
s\in
\mathcal{S}
=
\{\mathrm{early},\mathrm{ventral},\mathrm{dorsal}\}
\end{equation}
index the three visual encoding streams. For candidate $\mathbf{x}_k$, the
corresponding encoding model predicts
\begin{equation}
\widehat{\mathbf{y}}_{s,k}
=
E_s(\mathbf{x}_k),
\label{eq:encoding_prediction}
\end{equation}
where $E_s$ consists of the stream-specific visual feature extractor and the
ridge readout trained to predict activity in the corresponding cortical
regions.

The neural agreement for candidate $k$ is measured by the 
Pearson correlation
\begin{equation}
r_{s,k}
=
\operatorname{corr}
\left(
\mathbf{y}_s,
\widehat{\mathbf{y}}_{s,k}
\right).
\label{eq:weighted_corr}
\end{equation}
where $\mathbf{y}_s$ is the original neural activity.\\
For vectors $\mathbf{a}$ and $\mathbf{b}$ we define
\begin{equation}
\operatorname{corr}(\mathbf{a},\mathbf{b}) =
\frac{\sum_v (a_v-\bar a)(b_v-\bar b)}
{\sqrt{\sum_v (a_v-\bar a)^2}\,\sqrt{\sum_v (b_v-\bar b)^2}}
\end{equation}
where
\begin{equation}
\bar a = \frac{1}{V_s}\sum_v a_v
\qquad
\bar b = \frac{1}{V_s}\sum_v b_v
\end{equation}

Because the three streams produce scores on different scales, correlations
are standardized across the candidate set within each stimulus:
\begin{equation}
z_{s,k}
=
\frac{
r_{s,k}-\mu_{r,s}
}{
\sigma_{r,s}+\varepsilon
},
\qquad
\mu_{r,s}
=
\frac{1}{K}\sum_{j=1}^{K}r_{s,j},
\label{eq:score_z}
\end{equation}
with $\sigma_{r,s}$ the corresponding across-candidate (population) standard
deviation and $\varepsilon=10^{-9}$.
In the
candidate-set-size analysis in the main text, $\mu_{r,s}$, and $\sigma_{r,s}$ are calculated
once on the full pool of $100$ candidates and the selection is made within
each subset of $K$ candidates, to decrease noise and using the same reranker. This means that it measures how much the same selector gains choosing from a smaller subset of candidates.\\
The final neural compatibility score is the uniform average
\begin{equation}
S_k
=
S(\mathbf{y},\mathbf{x}_k;\mathbf{X}_{1:K})
=
\frac{1}{|\mathcal{S}|}
\sum_{s\in\mathcal{S}}z_{s,k},
\label{eq:combined_score}
\end{equation}
and the reconstruction is
\begin{equation}
\hat{\mathbf{x}}
=
\mathbf{x}_{k^\star},
\qquad
k^\star
=
\arg\max_{1\leq k\leq K}S_k.
\label{eq:final_selection}
\end{equation}

Importantly, $S_k$ is a \emph{neural compatibility score}, not a calibrated
likelihood $P(\mathbf{y}\mid\mathbf{x}_k)$. Because Equation \ref{eq:score_z} depends on the full
candidate set $\mathbf{X}_{1:K}$, the score assigned to one candidate is
defined relative to the other sampled candidates. This set dependence is why
we distinguish the implemented reranker from an exact Bayesian likelihood.

\subsection{Why semantic conditioning alone is insufficient}

The semantic proposal and neural reranker use complementary information from
the same brain response.

First, $\hat{\mathbf{c}}$ primarily constrains high-level content. It places
candidate generation in a region of the CogVideoX space compatible with the
semantic information decoded from fMRI, but does not uniquely determine scene
details. Consequently,
$q_\theta(\mathbf{x}\mid\mathbf{y})$ can contain visually distinct videos that
share similar semantics.

Second, conditioning on a deterministic estimate does not guarantee that all stimulus-relevant information in $\mathbf{y}$ is used by the generator. The encoding stage scores candidate realizations against $\mathbf{y}$ directly. This provides an alternative use of the measured response, but does not by itself prove that the conditioning map discards information. In this sense the two stages serve different
purposes:
\begin{equation}
\underbrace{
\mathbf{y}\rightarrow\hat{\mathbf{c}}
\rightarrow q_\theta(\mathbf{x}\mid\mathbf{y})
}_{\text{semantic search-space restriction}}
\qquad\text{and}\qquad
\underbrace{
(\mathbf{y},\mathbf{X}_{1:K})
\rightarrow S_1,\ldots,S_K
}_{\text{neural discrimination}}.
\end{equation}

The first stage answers \emph{which region of video space should be searched?}
The second asks \emph{which sampled realization is most consistent with the
measured neural response?} This separation is the central simplification of
our framework.

\subsection{Candidate-set size and the role of reranking}
\label{apd:sampling_theory}

Increasing the candidate-set size gives the reranker more hypotheses from the
semantic proposal, but it is important to distinguish what can be guaranteed
mathematically from what is observed empirically.

\paragraph{Proposition 2 (best-score monotonicity).}
Let
\begin{equation}
M_K
=
\max_{1\leq k\leq K}
s(\mathbf{y},\mathbf{X}_k),
\end{equation}
where $s(\mathbf{y},\mathbf{x})$ is a fixed pointwise compatibility function.
For nested candidate sets,
\begin{equation}
M_{K+1}\geq M_K.
\end{equation}

\paragraph{Proof.}
Adding one candidate enlarges the set over which the maximum is taken. Hence,
\begin{equation}
M_{K+1}
=
\max
\left\{
M_K,
s(\mathbf{y},\mathbf{X}_{K+1})
\right\}
\geq M_K.
\end{equation}
Therefore, the best attainable compatibility score cannot decrease when the
candidate set is enlarged. \hfill$\square$

If, in addition, the candidate scores
$s(\mathbf{y},\mathbf{X}_1),\ldots,s(\mathbf{y},\mathbf{X}_K)$
are i.i.d.\ with cumulative distribution function $F$, then
\begin{equation}
P(M_K\leq t)
=
P\!\left(
s(\mathbf{y},\mathbf{X}_1)\leq t,\ldots,
s(\mathbf{y},\mathbf{X}_K)\leq t
\right)
=
F(t)^K.
\end{equation}
Consequently,
\begin{equation}
P(M_K>t)=1-F(t)^K,
\end{equation}
which is non-decreasing in $K$ because $0\leq F(t)\leq1$.

Proposition~2 establishes monotonicity only for the maximum of a fixed
pointwise compatibility score as the candidate set grows; it does
\emph{not} imply that an external reconstruction metric such as SSIM or
identification accuracy must increase monotonically with $K$. Moreover, our implemented score $S(\mathbf{y},\mathbf{x}_k;\mathbf{X}_{1:K})$ is itself candidate-set dependent because of the within-set standardization. Therefore, Proposition~2 should be interpreted as providing theoretical intuition for candidate search rather than as a formal monotonicity guarantee for the implemented reranking procedure or its evaluation metrics.

The random-selection control admits a complementary result.

\paragraph{Lemma 1 (random-selection invariance).}
Conditioned on a neural observation $\mathbf{y}$, let
\[
\mathbf{X}_1,\ldots,\mathbf{X}_K
\overset{\mathrm{i.i.d.}}{\sim}
q_\theta(\mathbf{x}\mid\mathbf{y}),
\]
and let
$J\sim\operatorname{Uniform}\{1,\ldots,K\}$ independently of the candidates.
Then, for every $K$,
\begin{equation}
\mathbf{X}_J
\sim
q_\theta(\mathbf{x}\mid\mathbf{y}).
\end{equation}
Consequently, for any integrable reconstruction metric $Q$,
\begin{equation}
\mathbb{E}
\left[
Q(\mathbf{X}_J)
\mid \mathbf{y}
\right]
=
\mathbb{E}_{\mathbf{X}\sim q_\theta(\cdot\mid\mathbf{y})}
\left[
Q(\mathbf{X})
\right],
\end{equation}
which is independent of the candidate-set size $K$.

\paragraph{Proof.}
For any measurable set $A$,
\begin{align}
P(\mathbf{X}_J\in A\mid\mathbf{y})
&=
\sum_{k=1}^{K}
P(J=k)\,
P(\mathbf{X}_k\in A\mid\mathbf{y})
\\
&=
\frac{1}{K}
\sum_{k=1}^{K}
q_\theta(A\mid\mathbf{y})
\\
&=
q_\theta(A\mid\mathbf{y}).
\end{align}
Hence $\mathbf{X}_J$ has the same conditional distribution as any individual
sample from the proposal, independently of $K$. The equality of expectations
follows directly from equality in distribution. \hfill$\square$

Lemma~1 provides the theoretical basis for the random-selection control;
increasing $K$ alone does not improve the expected quality of a uniformly
selected candidate. Larger candidate pools become useful only when the
selection rule exploits information that distinguishes among their members.

Consistent with this prediction, the results in the main text show that brain-guided reranking generally improves as the candidate pool grows from $K=1$ to $K=100$, whereas random
selection remains approximately stable. The evaluation metrics are not
required to vary monotonically with $K$, and small fluctuations are indeed
observed; nevertheless, the overall trend supports the interpretation that
larger candidate pools provide the neural reranker with a richer set of
hypotheses from which to select.

\subsection{Relation to Bayesian decoding}
 Classical Bayes motivates two desirable
ingredients: a structured prior over plausible stimuli and a measure of
consistency with neural evidence. Our framework implements these ingredients
computationally as
\begin{equation}
\boxed{
\text{brain-conditioned proposal}
\quad+\quad
\text{encoding-based neural reranking}.
}
\end{equation}
The semantic decoder converts the broad generative model into an informative
proposal
$q_\theta(\mathbf{x}\mid\mathbf{y})$, while the encoding models provide a
relative compatibility criterion over the finite candidate set. This preserves
the central Bayesian intuition, combining generative structure with neural
evidence, without requiring either the proposal or the reranking score to be
interpreted as the exact posterior or likelihood.

\section{Quantitative Results of Ablation Studies}\label{quant_ablation}
\newcommand{\val}[2]{$#1 \pm #2$}
\begin{table}[H]
\centering
\small
\renewcommand{\arraystretch}{1.1}

\caption{Effect of the candidate-set size $K$ on reconstruction quality on CC2017, with encoding-model reranking versus random selection. Values are means across the three subjects, averaged over five nested random permutations of fixed candidate pools; $\pm$ denotes permutation standard deviation only. At $K=1$ reranking and random selection coincide. At $K=100$ both are invariant to these permutations, so their permutation standard deviation is zero.}
\label{tab:rerank-ablation-CC2017}
\tablebox{
\resizebox{0.92\textwidth}{!}{%
\begin{tabular}{lcccccc}
\toprule
$K$ &
\multicolumn{2}{c}{\textbf{Video 50-way}} &
\multicolumn{2}{c}{\textbf{Frame 50-way}} &
\multicolumn{2}{c}{\textbf{SSIM}} \\
\cmidrule(lr){2-3} \cmidrule(lr){4-5} \cmidrule(lr){6-7}
& rerank & random & rerank & random & rerank & random \\
\midrule
1   & $0.2359 \pm 0.0051$ & $0.2359 \pm 0.0051$ & $0.1872 \pm 0.0018$ & $0.1872 \pm 0.0018$ & $0.3951 \pm 0.0008$ & $0.3951 \pm 0.0008$ \\
20  & $0.2458 \pm 0.0019$ & $0.2373 \pm 0.0010$ & $0.2001 \pm 0.0007$ & $0.1887 \pm 0.0008$ & $0.3994 \pm 0.0009$ & $0.3946 \pm 0.0002$ \\
40  & $0.2476 \pm 0.0020$ & $0.2394 \pm 0.0013$ & $0.2033 \pm 0.0017$ & $0.1894 \pm 0.0009$ & $0.4004 \pm 0.0006$ & $0.3945 \pm 0.0002$ \\
60  & $0.2471 \pm 0.0029$ & $0.2387 \pm 0.0013$ & $0.2040 \pm 0.0033$ & $0.1894 \pm 0.0007$ & $0.4014 \pm 0.0006$ & $0.3945 \pm 0.0001$ \\
80  & $0.2466 \pm 0.0030$ & $0.2381 \pm 0.0010$ & $0.2042 \pm 0.0017$ & $0.1892 \pm 0.0004$ & $0.4016 \pm 0.0004$ & $0.3944 \pm 0.0001$ \\
100 & $0.2434$ & $0.2376$ & $0.2049$ & $0.1886$ & $0.4018$ & $0.3944$ \\
\bottomrule
\end{tabular}%
}
}
\end{table}

\begin{table}[H]
\centering
\small
\renewcommand{\arraystretch}{1.1}

\caption{Effect of the candidate-set size $K$ on reconstruction quality on BMD, with encoding-model reranking versus random selection. Values are means across the ten subjects, averaged over five nested random permutations of fixed candidate pools; $\pm$ denotes permutation standard deviation only. At $K=1$ reranking and random selection coincide. At $K=100$ both are invariant to these permutations, so their permutation standard deviation is zero.}
\label{tab:rerank-ablation-bmd}
\tablebox{
\resizebox{0.92\textwidth}{!}{%
\begin{tabular}{lcccccc}
\toprule
$K$ &
\multicolumn{2}{c}{\textbf{Video 50-way}} &
\multicolumn{2}{c}{\textbf{Frame 50-way}} &
\multicolumn{2}{c}{\textbf{SSIM}} \\
\cmidrule(lr){2-3} \cmidrule(lr){4-5} \cmidrule(lr){6-7}
& rerank & random & rerank & random & rerank & random \\
\midrule
1   & $0.3653 \pm 0.0203$ & $0.3653 \pm 0.0203$ & $0.3310 \pm 0.0038$ & $0.3310 \pm 0.0038$ & $0.4060 \pm 0.0046$ & $0.4060 \pm 0.0046$ \\
20  & $0.3748 \pm 0.0077$ & $0.3556 \pm 0.0092$ & $0.3512 \pm 0.0036$ & $0.3284 \pm 0.0049$ & $0.4087 \pm 0.0020$ & $0.4086 \pm 0.0006$ \\
40  & $0.3760 \pm 0.0034$ & $0.3588 \pm 0.0041$ & $0.3539 \pm 0.0061$ & $0.3276 \pm 0.0061$ & $0.4096 \pm 0.0013$ & $0.4085 \pm 0.0001$ \\
60  & $0.3807 \pm 0.0084$ & $0.3607 \pm 0.0054$ & $0.3567 \pm 0.0067$ & $0.3279 \pm 0.0041$ & $0.4108 \pm 0.0009$ & $0.4087 \pm 0.0003$ \\
80  & $0.3865 \pm 0.0041$ & $0.3624 \pm 0.0036$ & $0.3561 \pm 0.0036$ & $0.3270 \pm 0.0015$ & $0.4109 \pm 0.0008$ & $0.4086 \pm 0.0003$ \\
100 & $0.3920$ & $0.3634$ & $0.3535$ & $0.3257$ & $0.4116$ & $0.4086$ \\
\bottomrule
\end{tabular}%
}
}
\end{table}

\subsection{Further Experiments on the Contribution of the Reranker}\label{apd:shuffle}
For each test stimulus, we kept its 100 candidates, and therefore its decoded conditioning, fixed and replaced only the measured response used for scoring with that of another test stimulus. For BMD, responses were reassigned by a random permutation without fixed points; for CC2017, the reassigned response always came from a different test movie, to avoid using a temporally adjacent and thus similar clip. Results are averaged over five permutations and compared with random selection from the same pools and with the selection based on the reranker response (Table \ref{tab:shuffle}). Random selection is reported as its expected value, i.e. the mean of each metric over the $100$ candidates of each stimulus. With shuffled responses, semantic accuracy drops below random selection, indicating that the score is driven by the content of the measured response: when this response belongs to another stimulus, the selected candidates resemble that stimulus rather than the target. The reranker selection improves over the shuffled one on all semantic metrics. A generic component appears for SSIM on CC2017, where about half of the gain over random selection is also obtained with shuffled responses, and, to a smaller extent, for PSNR on BMD.\\
To test whether the measured response adds information beyond the semantics already decoded into the conditioning, we also reranked the same candidates without using the measured response, selecting for each stimulus the candidate most similar to its decoded conditioning $\hat{\mathbf{c}}$. To this end, a ridge regression estimated on training data only maps the averaged OpenCLIP bigG embeddings of video-frames to the T5 caption embedding averaged over tokens, and candidates are ranked by the cosine similarity between their projection and $\hat{\mathbf{c}}$ (averaged over tokens), both centered on the training mean. On BMD, conditioning-only selection improves over random selection on SSIM, with positive but non-significant differences on Video and Frame 50-way accuracy, whereas the measured response yields further gains on Video 50-way accuracy (borderline, $p=0.054$), Video 2-way accuracy and PSNR. On CC2017, conditioning-only selection improves over random selection  on Video 50-way and Frame 50-way accuracy, and the gains on Frame 50-way accuracy, SSIM and PSNR over conditioning-only selection require the measured response. Across both datasets, semantic accuracy follows the order shuffled $<$ random $<$ conditioning only $<$ reranked selection, indicating that the measured response contributes information beyond the decoded semantics, in particular for structural fidelity on CC2017.

\begin{table}[H]
\centering
\small
\renewcommand{\arraystretch}{1.1}

\caption{Shuffled-response and conditioning-only controls. Values are mean $\pm$ SEM across subjects (3 for CC2017, 10 for BMD). Random: expected value of a uniform choice among the $K=100$ candidates of the same pool (mean over candidates). Shuffled: candidates scored against the response to another test stimulus (mean of five permutations; for CC2017, always from a different movie). Conditioning only: candidate most similar to the decoded conditioning, without the measured response. Paired $t$-test across subjects of reranked versus shuffled: \textsuperscript{*}~$p<0.05$, \textsuperscript{**}~$p<0.01$, \textsuperscript{***}~$p<0.001$; reranked versus conditioning only: \textsuperscript{\dag}~$p<0.05$, \textsuperscript{\dag\dag}~$p<0.01$.}
\label{tab:shuffle}
\tablebox{
\resizebox{0.92\textwidth}{!}{%
\begin{tabular}{lcccc}
\toprule
\textbf{Selection} & Video 50-way & Frame 50-way & SSIM & PSNR \\
\midrule
\multicolumn{5}{c}{\textbf{CC2017}} \\
\midrule
Random            & \val{0.2376}{0.0121} & \val{0.1886}{0.0064} & \val{0.3944}{0.0019} & \val{9.022}{0.020} \\
Shuffled          & \val{0.2314}{0.0095} & \val{0.1799}{0.0061} & \val{0.3979}{0.0012} & \val{8.987}{0.016} \\
Conditioning only & \val{0.2412}{0.0090} & \val{0.1952}{0.0057} & \val{0.3947}{0.0018} & \val{8.999}{0.010} \\
Reranked          & \val{0.2434}{0.0162} & \val{0.2049}{0.0073}\textsuperscript{**\dag} & \val{0.4018}{0.0015}\textsuperscript{*\dag} & \val{9.146}{0.016}\textsuperscript{***\dag\dag} \\
\midrule
\multicolumn{5}{c}{\textbf{BMD}} \\
\midrule
Random            & \val{0.3634}{0.0082} & \val{0.3257}{0.0046} & \val{0.4086}{0.0012} & \val{9.851}{0.017} \\
Shuffled          & \val{0.3311}{0.0099} & \val{0.3130}{0.0037} & \val{0.4098}{0.0014} & \val{9.876}{0.019} \\
Conditioning only & \val{0.3715}{0.0105} & \val{0.3429}{0.0069} & \val{0.4116}{0.0013} & \val{9.880}{0.038} \\
Reranked          & \val{0.3920}{0.0115}\textsuperscript{***} & \val{0.3535}{0.0081}\textsuperscript{***} & \val{0.4116}{0.0016} & \val{9.987}{0.031}\textsuperscript{**\dag} \\
\bottomrule
\end{tabular}%
}
}
\end{table}

\section{Application of our Pipeline to NEURONS}\label{neurons_experiment}
\begin{figure}[H]
\centering
   {\includegraphics[width=\linewidth]{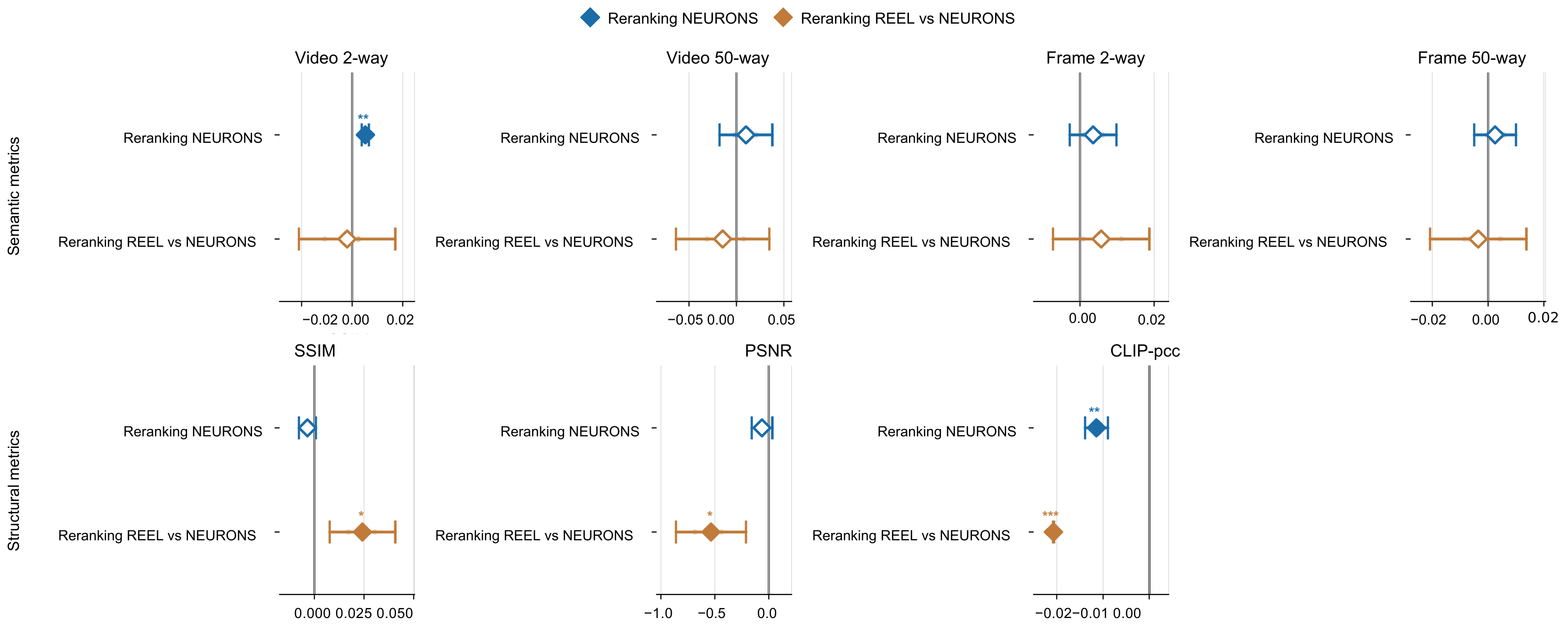}}
  \caption{Paired subject-level differences in mean performance with respect to the NEURONS baseline on CC2017,
   for our reranking method applied to the NEURONS candidate pool (blue) and for our
   proposed pipeline (orange). Each panel reports one metric on its own scale (top row:
   semantic metrics; bottom row: structural metrics); the vertical line at zero
   corresponds to the NEURONS baseline. Diamonds denote the mean difference across the
   three subjects and whiskers its $95$\% confidence interval (paired $t$-test,
   $\mathrm{df}=2$); filled markers indicate intervals that exclude zero, and faint dots
   show the individual subjects. \textsuperscript{*}~$p<0.05$, \textsuperscript{**}~$p<0.01$, \textsuperscript{***}~$p<0.001$.}
 \label{fig:stats}
\end{figure}

We reproduced the NEURONS pipeline to evaluate our reranking procedure on candidates generated by their framework on CC2017. Using the released configuration (AnimateDiff v3 on Stable Diffusion 1.5 with the Realistic Vision V6.0 B1 checkpoint, v3 motion module and adapter, SparseCtrl RGB conditioning on the reconstructed keyframe, blurry-video and predicted-caption conditioning; $25$ inference steps, guidance scale $8.5$, $16$ frames at $256\times256$), we generated $K=100$ candidates per test stimulus with seeds $42$--$141$, and applied our reranking with the same encoding models and selection rule as in our pipeline. The NEURONS baseline is a single generation with the default seed of the configuration (seed $0$, not included among the candidates), without any selection or aggregation; the comparison therefore uses the same generator and settings, and differs only in the generation budget and in the selection. Our experiments reveal a limitation when applying our approach to this candidate pool, which may arise from the stronger guidance employed during the generation process. Specifically, NEURONS uses the generated key-object masks to condition both image and blurry-video generation, while the generated scene description is provided as a textual prompt to the text-to-video diffusion model. As a result, the generated candidates exhibit smaller variations: the mean pairwise cosine distance between the OpenCLIP ViT-bigG embeddings of candidates of the same stimulus is $0.13$ for NEURONS against $0.35$ for our pipeline (consistent across the three subjects), and the within-stimulus standard deviation of the evaluation metrics across candidates is $35$--$40$\% lower, making candidates more difficult to discriminate based on subtle differences in their predicted neural responses. Nevertheless, as reported in Table \ref{tab:main-comparison}, our reranking procedure still provides improvements on semantic metrics, although small gains on frame-level semantic metrics and slight, non-significant decreases in SSIM and PSNR than those observed with our proposed pipeline.\\
Figure \ref{fig:stats} reports the paired differences with respect to the NEURONS baseline, together with their $95$\% confidence intervals across subjects. When applied to the NEURONS candidate pool (blue), our reranking significantly improves Video 2-way accuracy, while the other semantic metrics show positive but non-significant differences. This gain comes with a small but significant decrease in CLIP-pcc, whereas SSIM and PSNR are not significantly affected. For REEL versus NEURONS (orange), the confidence intervals for semantic-accuracy differences include zero; with three subjects, this does not establish equivalence. REEL has higher mean SSIM and lower mean PSNR and CLIP-pcc. The experiment does not isolate which pipeline differences account for these trade-offs.

\begin{table}[H]
\centering
\small
\renewcommand{\arraystretch}{1.1}

\caption{Comparison between the baseline and our method, averaged over the three subjects. Values are mean $\pm$ SEM across subjects. The best value per metric is in bold. \textsuperscript{*}~$p<0.05$, \textsuperscript{**}~$p<0.01$, \textsuperscript{***}~$p<0.001$ paired $t$-test across subjects ($\mathrm{df}=2$).}
\label{tab:main-comparison}
\tablebox{
\begin{tabular}{lcc}
\toprule
\textbf{Metric} & \textbf{NEURONS} & \textbf{NEURONS + reranking} \\
\midrule
Video 2-way  & \val{$0.8575$}{$0.0039$} & $\mathbf{0.8627}\pm0.0037$\textsuperscript{**} \\
Video 50-way & \val{$0.2579$}{$0.0068$} & $\mathbf{0.2680}\pm0.0073$ \\
Frame 2-way  & \val{$0.8083$}{$0.0054$} & $\mathbf{0.8118} \pm0.0067$ \\
Frame 50-way & \val{$0.2084$}{$0.0055$} & $\mathbf{0.2110} \pm0.0042$ \\
SSIM         & $\mathbf{0.3776} \pm0.0038$ & \val{$0.3740$}{$0.0037$} \\
PSNR         & $\mathbf{9.682}\pm0.087$ & \val{$9.618$}{$0.089$} \\
CLIP-pcc     & $\mathbf{0.9367}\pm0.0005$ & \val{$0.9253$}{$0.0010$}\textsuperscript{**} \\
\bottomrule
\end{tabular}
}
\end{table}

\newpage

\section{Qualitative Results}\label{apd:second}
We show manually selected examples spanning higher, intermediate and lower reconstruction quality for CC2017 and BMD. These examples are meant to be illustrative rather than an estimate of typical performance, since they were selected by visual inspection.

\subsection{CC2017}

\begin{figure}[H]
  \centering
  \includegraphics[width=\linewidth]{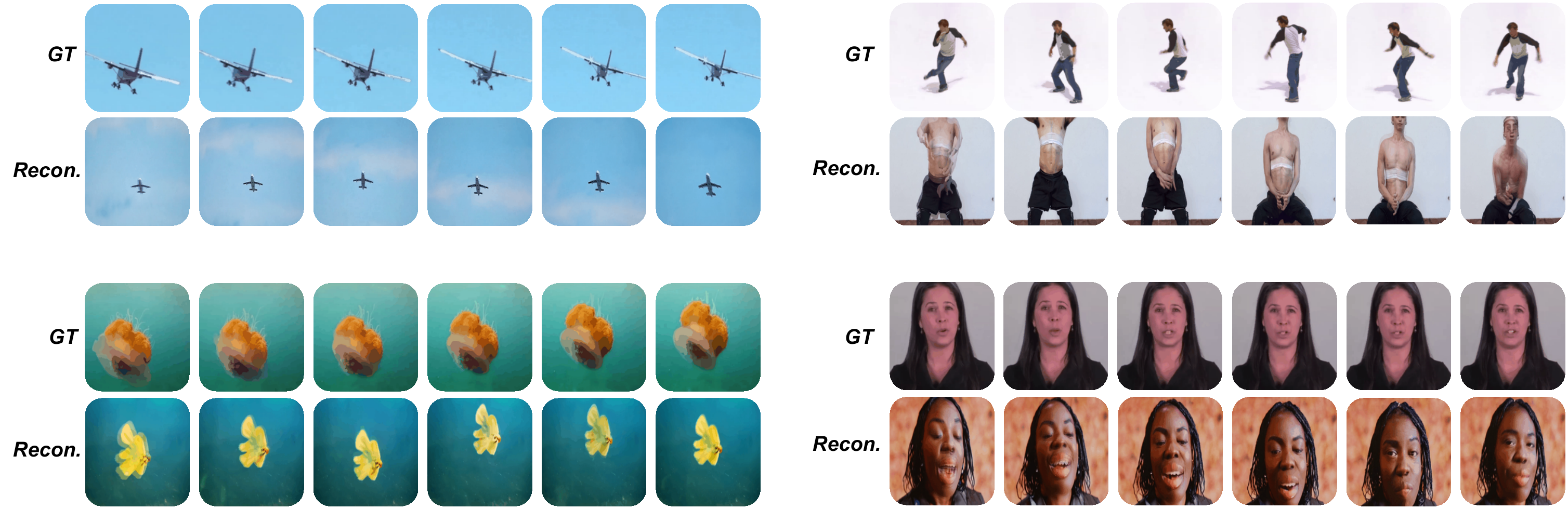}
  \caption{\textbf{Best reconstructions on CC2017.} These frames show a strong correspondence between the original scene shown in the video and our reconstructions, capturing the overall layout, color palette and coarse semantic category.}
\end{figure}

\begin{figure}[H]
  \centering
  \includegraphics[width=\linewidth]{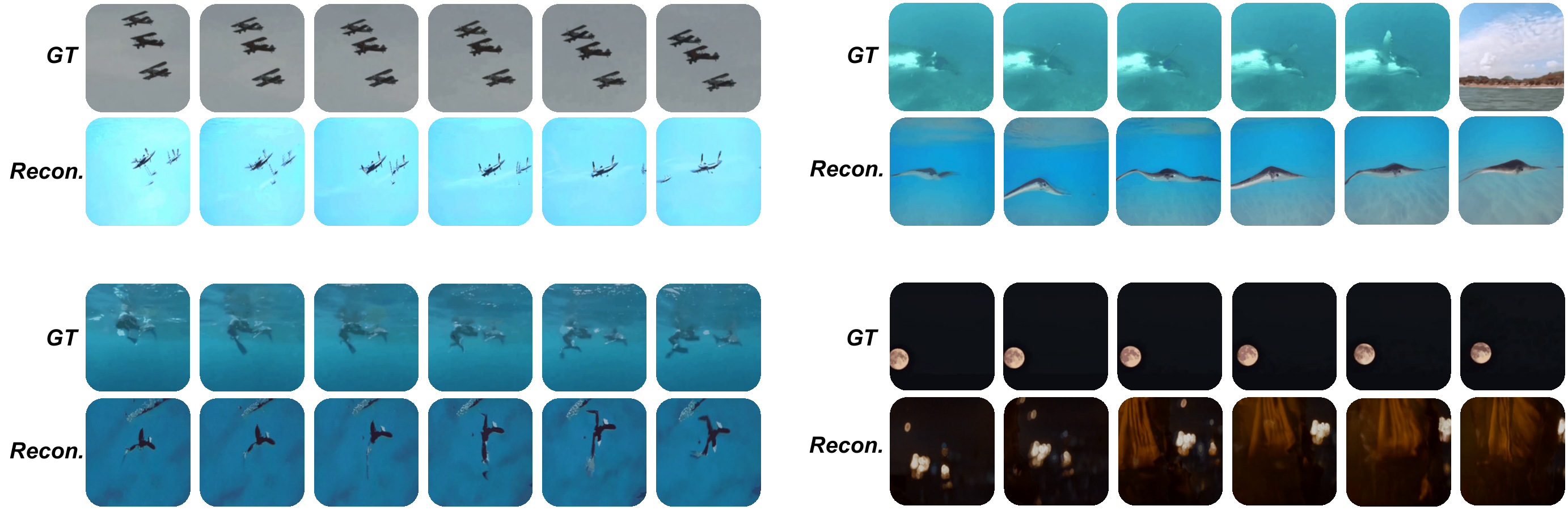}
  \caption{\textbf{Intermediate reconstructions on CC2017.} The reconstructions correctly reproduce the subjects and the environment of the original videos, but show limitations with out-of-distribution elements, such as the moon appearing in the sky, and with scene transitions. Even in these cases, the general setting, such as the presence of a dark sky, is preserved.}
  \label{fig:medie}
\end{figure}

\begin{figure}[H]
  \centering
  \includegraphics[width=\linewidth]{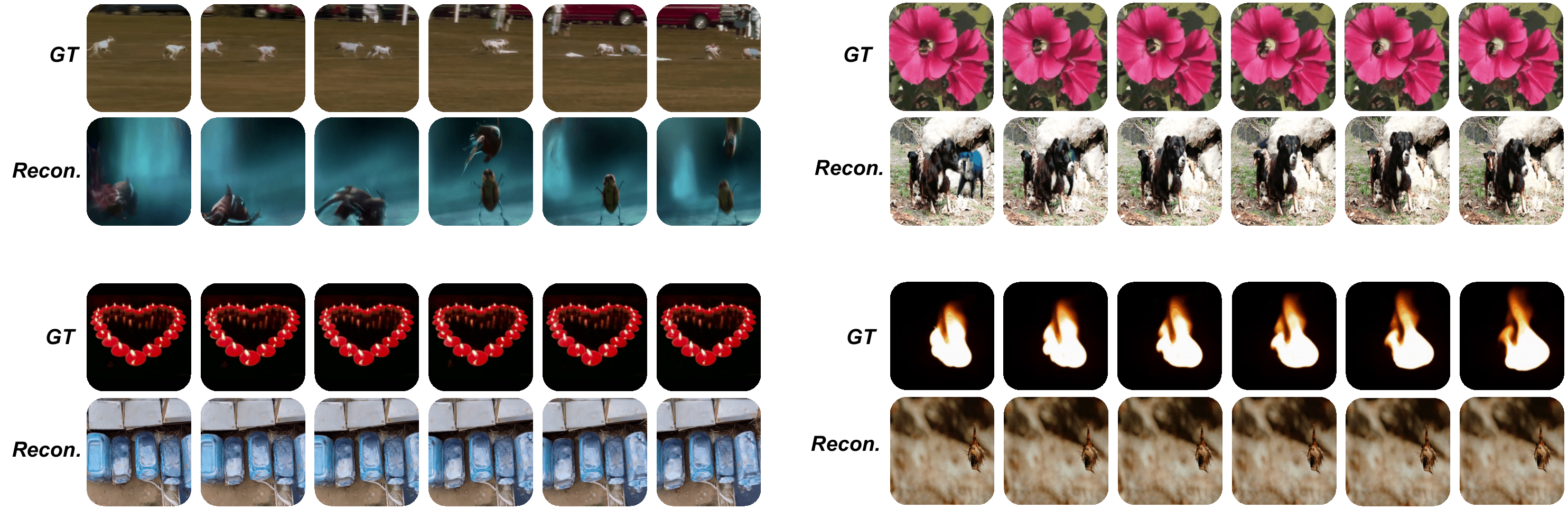}
  \caption{\textbf{Worst reconstructions on CC2017.} The reconstructions fail to reproduce the subjects and the environment of the original videos, mostly because these stimuli are out of distribution for our model. Nevertheless, coarse properties such as the presence of an action or the number of subjects in the scene still appear to be captured.}
  \label{fig:cattive}
\end{figure}

\clearpage
\subsection{BMD}

\begin{figure}[H]
  \centering
  \includegraphics[width=\linewidth]{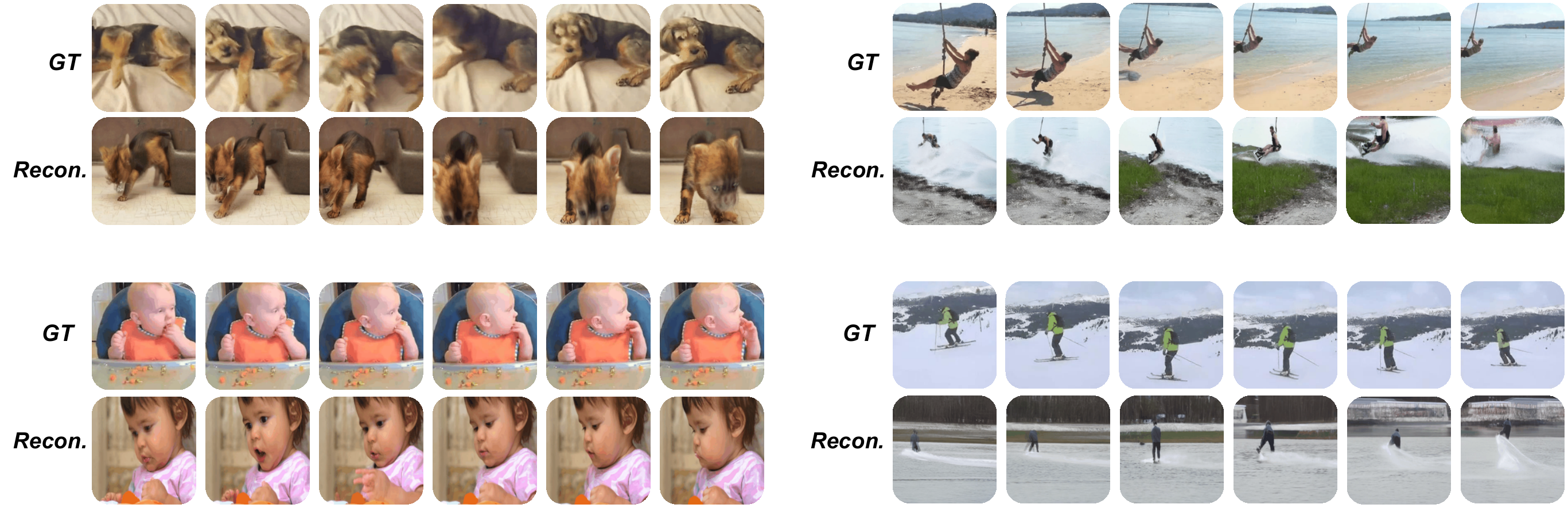}
  \caption{\textbf{Best reconstructions on BMD.} These frames show a strong correspondence between the original scene displayed on the video and our reconstructions, capturing the overall layout, color palette and coarse semantic category.}
\end{figure}

\begin{figure}[H]
  \centering
  \includegraphics[width=\linewidth]{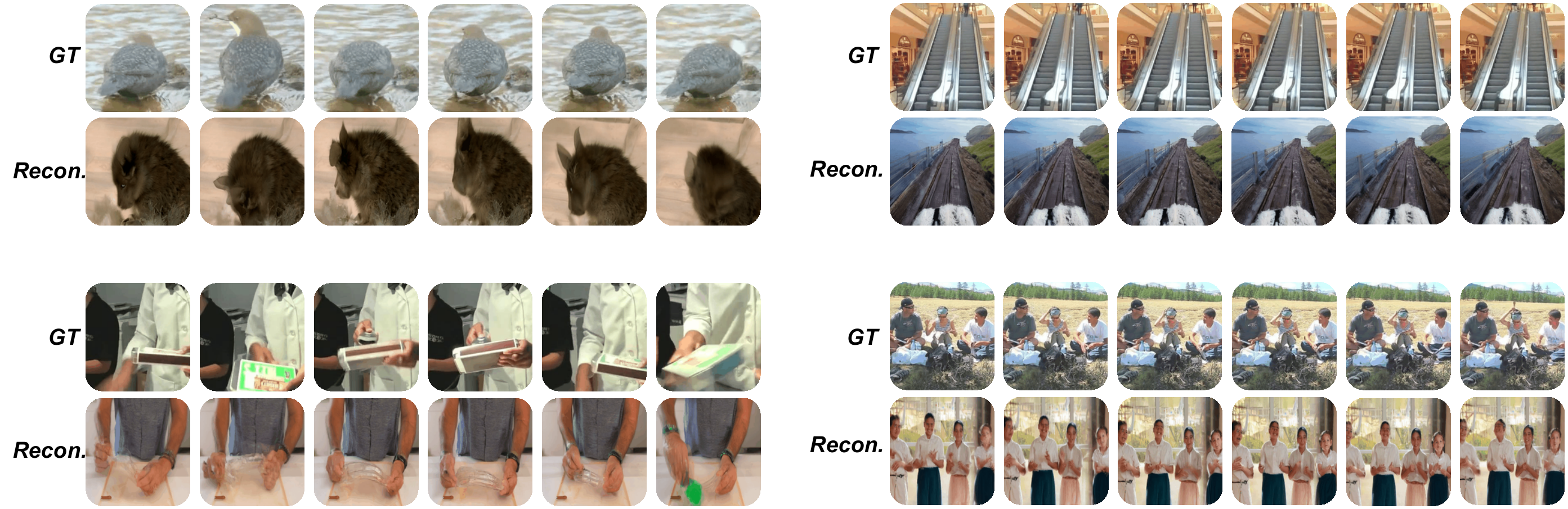}
  \caption{\textbf{Intermediate reconstructions on BMD.} The reconstructions correctly reproduce the subjects and the environment of the original videos, but show limitations. However, even in these cases, the general setting and the number of subjects resemble the corresponding ground truth, is preserved. 
  }
  \label{fig:medie_bold}
\end{figure}

\begin{figure}[H]
  \centering
  \includegraphics[width=\linewidth]{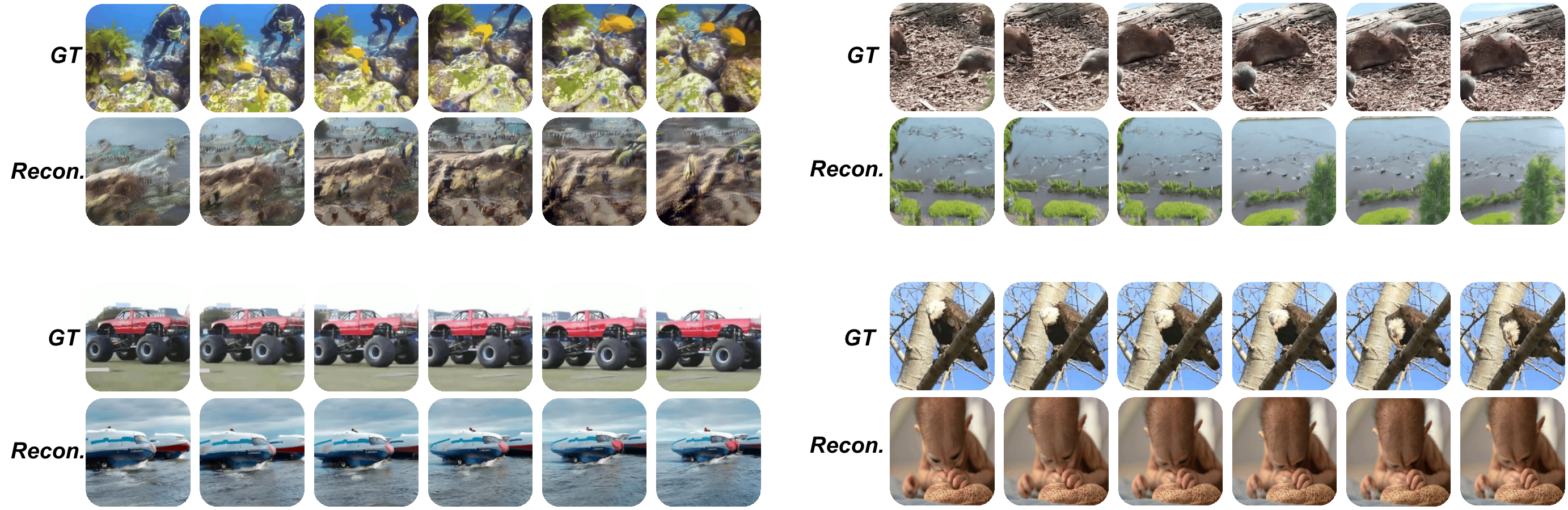}
  \caption{\textbf{Worst reconstructions on BMD.} The reconstructions fail to reproduce the subjects and the environment of the original videos, mostly because these stimuli are out of distribution for our model. 
  }
  \label{fig:cattive_bold}
\end{figure}

\end{document}